\documentclass[10pt,journal,compsoc]{IEEEtran}

\ifCLASSOPTIONcompsoc
\usepackage[nocompress]{cite}
\else
\usepackage{cite}
\fi

\usepackage{times}

\usepackage{graphicx,epsfig,subcaption}

\usepackage{amsmath,amssymb}
\usepackage{booktabs,multirow,makecell}

\usepackage{float}
\usepackage[norelsize, linesnumbered, ruled, lined, boxed, commentsnumbered]{algorithm2e}
\usepackage{xcolor}
\usepackage[pagebackref=true,colorlinks,citecolor=blue!90!black,linkcolor=red!90!black,bookmarks=true]{hyperref}

\usepackage{xspace}

\usepackage{ragged2e}
\newcommand{\justfy}{\leftskip=0pt \rightskip=0pt plus 0cm}

\def\eg{e.g.\@\xspace}
\def\ie{i.e.\@\xspace}
\def\etal{et~al.\@\xspace}
\def\etc{etc\@\xspace}

\newcommand{\var}{\sigma^{2}}
\newcommand{\vari}{\sigma^{2}_{i}}

\newcommand{\G}{G}
\newcommand{\D}{D}
\newcommand\E{\mathbb{E}}

\newcommand{\x}{\mathbf{x}}
\newcommand{\z}{\mathbf{z}}
\newcommand{\xt}{\tilde{\mathbf{x}}}
\newcommand{\xa}{\mathbf{x}_{1}}
\newcommand{\xb}{\mathbf{x}_{2}}
\newcommand{\xii}{\mathbf{x}_{i}}
\newcommand{\xjj}{\mathbf{x}_{j}}
\newcommand{\xmixup}{\mathbf{x}'}
\newcommand{\xremix}{\mathbf{x}^{*}}

\newcommand{\pdata}{p_{\textrm{data}}}
\newcommand{\ppdata}{q_{\sigma}}

\newcommand{\weight}{\lambda}

\newcommand{\mixup}{f^{\textrm{mixup}}}

\newcommand{\scoremix}{f^{\textrm{scoremix}}}

\newcommand{\loss}[1]{\mathcal{L}^{#1}}

\newcommand{\ltwo}[1]{\|{#1}\|_{2}^{2}}

\newcommand\stein{\mathbb{S}}
\newcommand\snet{S_{\theta}}
\newcommand\sneto{S_{\theta^{*}}}
\newcommand{\argmin}[1]{\underset{#1}{\arg \min}}

\begin{document}
\title{ScoreMix: A Scalable Augmentation Strategy for Training GANs with Limited Data}

%
\author{
        Jie~Cao, 
        Mandi~Luo, 
        Junchi~Yu,
        Ming-Hsuan~Yang, 
        and~Ran~He
\IEEEcompsocitemizethanks{
\IEEEcompsocthanksitem Jie Cao, Mandi Luo, Junchi Yu and Ran He are with CRIPAC \& NLPR, Institute of Automation, Chinese Academy of Sciences, Beijing 100190, and University of Chinese Academy of Sciences, Beijing 101408. E-mail: jie.cao@cripac.ia.ac.cn, \{luomandi2019, yujunchi2019\}@ia.ac.cn, rhe@nlpr.ia.ac.cn.
\IEEEcompsocthanksitem Ming-Hsuan Yang is with Department of Computer
Science and Engineering at University of California, Merced. E-mail: mhyang@ucmerced.edu.
\IEEEcompsocthanksitem Corresponding author: Ran He.
}
}

\markboth{~}%
{Shell \MakeLowercase{\textit{et al.}}: Bare Advanced Demo of IEEEtran.cls for IEEE Computer Society Journals}

\linespread{1}
{ 
\IEEEtitleabstractindextext{%
\begin{abstract}
\justfy
Generative Adversarial Networks (GANs) typically suffer from overfitting when limited training data is available.
To facilitate GAN training, current methods propose to use data-specific augmentation techniques.
Despite the effectiveness, it is difficult for these methods to scale to practical applications.
In this work, we present ScoreMix, a novel and scalable data augmentation approach for various image synthesis tasks.
We first produce augmented samples using the convex combinations of the real samples.
Then, we optimize the augmented samples by minimizing the norms of the data scores, \ie, the gradients of the log-density functions.
This procedure enforces the augmented samples close to the data manifold.
To estimate the scores, we train a deep estimation network with multi-scale score matching.
For different image synthesis tasks, we train the score estimation network using different data.
We do not require the tuning of the hyperparameters or modifications to the network architecture. 
The ScoreMix method effectively increases the diversity of data and reduces the overfitting problem.
Moreover, it can be easily incorporated into existing GAN models with minor modifications. 
Experimental results on numerous tasks demonstrate that GAN models equipped with the ScoreMix method achieve significant improvements.
\end{abstract}

\begin{IEEEkeywords}
Generative Adversarial Networks, Image Synthesis, Data Augmentation, Few-Shot Image-to-Image Translation
\end{IEEEkeywords}
}

\maketitle
\IEEEdisplaynontitleabstractindextext
\IEEEpeerreviewmaketitle
}

\IEEEraisesectionheading{\section{Introduction}}
\IEEEPARstart{I}{n} recent years, Generative Adversarial Networks (GANs)~\cite{goodfellow2014generative} have shown much progress in numerous image synthesis tasks. 
However, current GAN-based methods~\cite{karras2017progressive,karras2020analyzing,brock_large_2019} heavily rely on vast quantities of training data. 
Large-scale training datasets, e.g., the CelebA~\cite{liu2015deep} and FFHQ~\cite{karras2019style} datasets, are necessary for these methods to achieve state-of-the-art results.
When the amount of data is limited, recent findings~\cite{karras_training_2020,zhao2020differentiable} reveal that GANs easily overfit the training set.
This issue leads to drastically degraded results in scenarios where collecting sufficient training data is infeasible.
Hence, improving the generalization of GANs to the data-limited regime is practically important to the image synthesis tasks.

To tackle the overfitting problem, some recent approaches develop data augmentation schemes~\cite{zhang2019consistency,karras_training_2020,zhao2020differentiable,tran_data_2021} tailored for GANs.
These schemes use a set of hand-crafted image processing operations (\eg, random cropping, scaling, and color jittering) to augment the training samples. 
They combine these operations with adaptive~\cite{karras_training_2020} or automatic~\cite{zhao2020differentiable} strategies.
For some specific tasks~\cite{zhao2020differentiable}, these methods can train GANs effectively with hundreds of samples, allowing decent performances.

However, the current methods have some intrinsic limitations: 
First, these methods are difficult to generalize across datasets. 
The reason is that domain knowledge is integrated into the hand-crafted augmentation operations of these methods. 
Such knowledge is valid only for some specific data. 
For instance, the color distortion strategies for horse images are not suitable for zebra images. 
Second, these methods include a large number of hyperparameters that require fine-tuning on specific training data. 
Note that unsuitable choices of the hyperparameters are even harmful to model training~\cite{karras_training_2020,chen_simple_2020}.
Fine-tuning these hyperparameters, however, is prohibitively expensive due to the large search space.
To reduce the computational burden, current methods assign fixed values to most of the hyperparameters based on their experience.
Due to these issues, training GANs with limited data in real-world scenarios remains challenging.

\begin{figure}[t]
\begin{center}
    \includegraphics[width=1.0\linewidth]{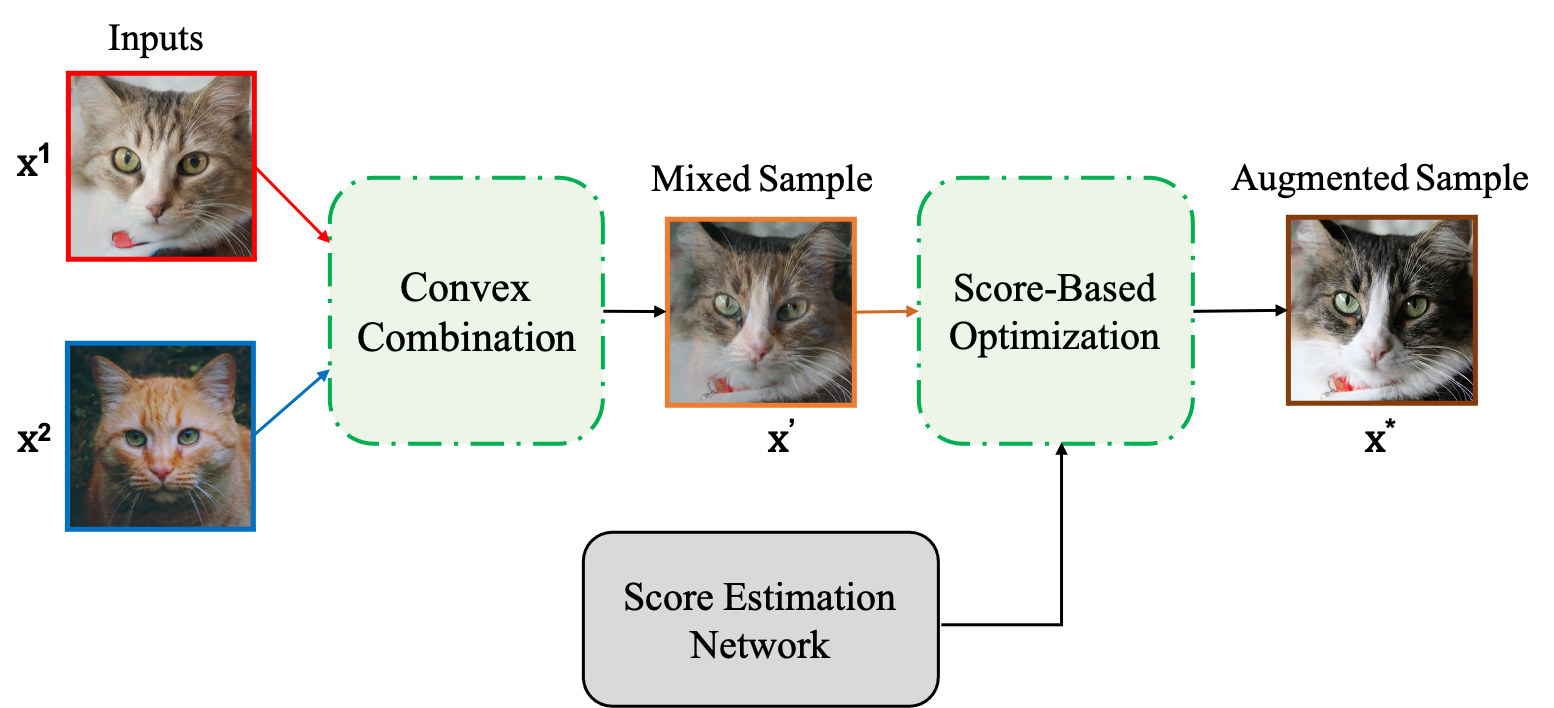}
\end{center}
\caption{
Illustration of the proposed data augmentation method.
Given two data points $\xa$ and $\xb$, we first take the convex combination of the two poinst as the mixed sample $\xmixup$.
Then, we optimize the mixed sample with a score estimation network.
The optimization strategy enfores the mixed sample to move to the high-density regions of the data distribution.
}
\label{fig:1}
\end{figure}

To facilitate training GANs with limited data, we propose a mixing-based augmentation strategy named ScoreMix.
As illustrated in Figure~\ref{fig:1}, given a data distribution $\pdata$, we first take a convex combination of two training samples as the mixed sample~$\xmixup$. 
Then, we optimize $\xmixup$ by minimizing the L2 norm of its Stein score~\cite{stein_bound_1972}.
The Stein score of a sample, denoted as $\nabla \log \pdata(\x)$, is the gradients of the logarithmic distribution function with respect to data.
During the optimization process,  the mixed sample $\xmixup$ gradually moves towards the manifold of data distribution.
 Under mild assumptions, for the solution $\xremix$, the density probability $\pdata(\xremix)$ has the maximal value in some local neighborhood around the mixed sample $\xmixup$.
 This means the augmented sample $\xremix$ has the maximum likelihood that belongs to the data distribution. 

To estimate the Stein scores of samples, we build a deep estimation network using the denoising score matching method~\cite{vincent_connection_2011}. 
During the training stage, the estimation network learns to predict the Stein scores of the samples perturbed by multi-scale random noises.
The minimal noise level is negligible, so that the network can accurately predict the Stein scores of the real samples.
We do not introduce any domain knowledge in the training process. Hence, the proposed method is general-purpose in nature. 
For different image synthesis tasks, we train the score estimation network only with the given training data. 
Moreover, the proposed method does not require any modifications to the hyperparameters or network architectures.

The ScoreMix method can be incorporated into the GAN models easily. 
We only need to modify how these methods feed training data. 
In addition, our approach is fully compatible with existing augmentation methods~\cite{zhao2020differentiable,lecun_gradient-based_1998,krizhevsky_imagenet_2012,devries2017improved,tseng_regularizing_2021}.

We evaluate the proposed method on several image synthesis tasks.
Extensive experiments are conducted on the FFHQ dataset~\cite{karras_style-based_2019}, the ImageNet dataset~\cite{krizhevsky_imagenet_2012}, MetFace dataset~\cite{karras_training_2020}, and AFHQ dataset~\cite{choi2020stargan}.
The results demonstrate that the GAN models equipped with the ScoreMix method achieve significant improvements.
Notably, we apply the proposed method to one-shot image-to-image translation tasks, where a single training sample is available for each image domain.
We also compare against the state-of-the-art mixing-based augmentation methods~\cite{zhang2017mixup,yun_cutmix_2019,hendrycks_augmix_2020}, and show that the proposed ScoreMix method performs favorably regarding both quantitative evaluations and human evaluations.

The main contributions are summarized as follows:

\begin{itemize}
\item
\textcolor{black}{	 	
   We present a score-based data augmentation method scalable to various image synthesis tasks.
   The augmented samples diversify training data without perturbing the data distribution.
   Our method reduces the overfitting problem, facilitating the training of GANs with limited data.
   }
\item 
   We achieve significant improvements in multiple image synthesis tasks.
   We produce plausible results with a small amount of training data.
   In addition, the proposed method works decently in one-shot image synthesis.  
\end{itemize}

This paper is an extension of our previous conference paper~\cite{cao_remix_2021}. 
The major improvements over the preliminary one are in three-folds: 
(1) The proposed data augmentation method can be applied to general image synthesis tasks. 
The previous approach is effective only for the image-to-image translation tasks.
(2) This work proposes to produce augmented samples with score-based optimization. 
Moreover, we build a score estimation network based on multi-scale denoising score matching.
This approach ensures that the augmented samples are close to the data manifold.
In contrast, the previous approach produces augmented samples by linearly interpolating the real training samples.
(3) Our approach does not rely on specific designs of the underlying GAN model. 
Thus, our approach facilitates the training of various GANs.
The previous approach, however, requires a predefined content-preserving loss for augmentation.

\section{Related Work}

\subsection{Data Augmentation.}
\label{sec:21}
Numerous methods have been developed to increase the amount of data for training deep learning models without overfitting. 
To name a few, Lecun~\etal~\cite{lecun_gradient-based_1998} use the affine transformations, including translation, scaling, and shearing, in handwritten character recognition; 
Bengio~\etal~\cite{bengio_deep_2011} introduce the transformations that degrade images (\eg, Gaussian noise, salt and pepper noise, and motion blur) for augmentation; 
Krizhevsky~\etal~\cite{krizhevsky_imagenet_2012} apply random cropping, horizontal flipping, and color jittering to train deep networks. 
Using these content-preserving transformations has become a routine data pre-processing step. 

The strategies mentioned above can effectively diversify the training data, but the hyperparameters require data-specific tuning, which is non-trivial in practice.
Note that an unsuitable augmentation setting even hurts model performance~\cite{karras_training_2020,chen_simple_2020}. 
\textcolor{black}{
Some methods~\cite{cubuk_autoaugment_2019,lim_fast_2019,cubuk_randaugment_2020,zhang_adversarial_2020} propose to automatically search the optimal hyperparameters for generic augmentation operations.
However, these methods are prohibitively expensive in many practical settings.
}

The mixing-based augmentation methods~\cite{devries2017dataset,zhang2017mixup,beckham2019adversarial,yun_cutmix_2019,hendrycks_augmix_2020,dabouei_supermix_2021,kim_puzzle_2020,uddin_saliencymix_2021,guo_mixup_2019,kalantidis_hard_2020} use the combinations of real samples to generate virtual training samples. 
These methods can regularize the models from the overfitting problem and improve generalization. The mixup method~\cite{zhang2017mixup} is one of the most widely used approaches, which uses linear interpolations for augmentation. 
This method is easy to implement and requires neglectable computation. However, the augmented samples look unnatural and exhibit visible artifacts, which may confuse the training model. 
Recent approaches have proposed various improved mixing-based methods. 
Guo~\etal~\cite{guo_mixup_2019} present an adaptive mixing policy to reduce the generation of degraded data. The CutMix method~\cite{yun_cutmix_2019} augments samples by cutting and pasting image patches within a data batch. 
The PuzzleMix method~\cite{kim_puzzle_2020} proposes to search for an optimal mask, which reveals the most salient objects in the real images, for data mixing. 
Dabouei~\etal~\cite{dabouei_supermix_2021} present the SuperMix method, which uses the knowledge from a teacher network for data augmentation. 
The AugMix method~\cite{hendrycks_augmix_2020} proposes a chain of augmentation operations and enforces that the augmented image is similar to the original one. Though the existing mixing-based methods have demonstrated notable improvements, most methods generate augmented samples that drift off the data manifold. Using such augmented samples might lead to degraded image quality for the image synthesis tasks. To address this issue, we propose to generate the mixed data with the maximum likelihood that belongs to the real data distribution. 

\subsection{Data-Efficient GANs}

GANs~\cite{goodfellow2014generative} are one of the most popular generative models. This method has been successfully applied to many computer vision applications. 
In particular, the GAN-based methods~\cite{gulrajani_improved_2017,arjovsky_wasserstein_2017,lucic_are_2018,mescheder_which_2018} significantly improve the visual quality and diversity of the synthesized images.
The state-of-the-art GAN models~\cite{karras_style-based_2019,karras2020analyzing,brock_large_2019} yield impressive synthesis results on large-scale datasets.
However, a fundamental challenge exists: the generalization ability and training stability of GANs are limited, especially in the data-limited regime.

These issues can be mitigated by collecting a large amount of data.
But this can be prohibitively expensive or implausible. 
Hence, some efforts~\cite{zhang2019consistency,tran2020towards,zhao2020image,karras_training_2020,zhao2020differentiable,tran_data_2021} have recently been made to augment training data for GANs. 
Karras~\etal~\cite{karras_training_2020} propose an adaptive data augmentation scheme for the discriminator.
During training, this scheme controls the augmentation strength based on the degree of overfitting.
The DiffAug method~\cite{zhao2020differentiable} introduces differentiable augmentation operations, which regularize GANs and improve the training stability.
Tran~\etal~\cite{tran_data_2021} propose an augmentation framework with invertible transformations.
They also provide theoretical analysis to prove this approach preserves the Jensen-Shannon divergence of original GANs.
These methods reduce the overfitting problem and render some plausible results.
However, they use data-specific augmentation operations.
Recent studies~\cite{karras_training_2020,zhao2020differentiable,chen_simple_2020} show that these methods are sensitive to the choices of datasets, training methods, and network architectures.
In this work, we propose a scalable data augmentation method for GAN training.
The proposed augmentation method does not require data-specific tuning of hyperparameters.
For different image synthesis tasks, we retrain the score estimation network, which guides the augmentation process.

The adaption-based approaches~\cite{liu2019few,saito2020coco,gu_ladn_2019,zakharov_few-shot_2019,wang_example-guided_2019,wang_few-shot_2019,han_viton_2018} train GANs with external datasets as an alternative to the target dataset.
Liu \etal~\cite{liu2019few} first learn a semantically related generation model and then adapt it to the target domain.
To deal with the content loss problem, Saito \etal~\cite{saito2020coco} propose the content-conditioned style encoder.
Moreover, several adaption-based methods focus on generating specific images, including faces~\cite{gu_ladn_2019,zakharov_few-shot_2019}, scenes~\cite{wang_example-guided_2019}, human bodies~\cite{wang_few-shot_2019,han_viton_2018}, \etc.
Despite the effectiveness, these approaches require additional image collection.
They also assume that the external dataset and target dataset share a great similarity.
In this paper, the proposed method trains GANs without any additional training data.

\textcolor{black}{
A recent work by Tseng \etal~\cite{tseng_regularizing_2021} proposes the \textbf{L}e\textbf{C}am-divergence (LC) regularization to improve the generalization of GANs trained with limited data.
They impose the LC regularization on the training objective of the discriminator.
The strategy can be easily added to the GAN training pipeline.
While in this work, we propose to improve the training of GANs by augmenting the data, which is orthogonal to the LC regularization.
We show that applying the LC regularization to the training process of our method brings additional improvements (later in our experiments).
}

Some recent approaches~\cite{shocher_zero-shot_2018,zhou_non-stationary_2018,shocher_ingan_2019,shocher_ingan_2019,shaham_singan_2019,hinz_improved_2021,park_contrastive_2020} investigate the one-shot image synthesis task (a.k.a, single image generation). 
These methods learn a generation model from one sample for each image domain. 
The one-shot generation methods typically compute the internal statistics of patches within the training sample for model training. 
Several methods use domain knowledge and solve specific one-shot tasks, \eg, texture synthesis~\cite{zhou_non-stationary_2018} and image super-resolution~\cite{shocher_zero-shot_2018}. 
Towards generalized one-shot image synthesis, Shocher~\etal~\cite{shocher_ingan_2019} present InGAN, a unified model for various tasks and data types.
Shaham~\etal~\cite{shaham_singan_2019} train a pyramid of GANs using a multi-scale pipeline. 
They produce synthesized results with a coarse-to-fine model inference. 
Hinz~\etal~\cite{hinz_improved_2021} propose a rescaling approach to reduce training time and a refinement method to improve image quality. 
To encourage content preservation, the SinCUT method~\cite{park_contrastive_2020} introduces a path-based contrastive learning scheme. 
In this work, the proposed data augmentation method is powerfully complementary to these approaches. 
Using the proposed method renders better results in the one-shot image synthesis tasks in our experiments.

\subsection{Score-Based Generative Model}
The score-based generative models~\cite{song_generative_2019,song_improved_2020,song_score-based_2021} represent data distributions through scores, the vector fields which reflect the gradients of the likelihood of data.
Similar to GANs, these models can produce high-quality images.
They have proven highly effective in multiple image synthesis tasks.
Song and Ermon~\cite{song_generative_2019} propose to train a neural network to estimate the scores of data. Then, they sample synthesized images using Langevin dynamics.
Following this line of research, Song and Ermon~\cite{song_improved_2020} later improve the sampling process with a set of techniques that effectively estimate the noise scales from training data.
Ho \etal~\cite{ho_denoising_2020} model the image sampling procedure as a reverse diffusion process.
Then, Song~\etal~\cite{song_score-based_2021} integrate previous methods into a unified framework with stochastic differential equations.
In this work, we train a deep score estimation model for data augmentation. 
Note that in the sampling process, we use the mixed image as initialization and produce the synthesized image with gradient-based optimization.
These differences distinguish our approach from the discussed methods.

\begin{figure*}
\centering
     \begin{subfigure}[b]{0.35\textwidth}
         \centering
         \includegraphics[width=\textwidth]{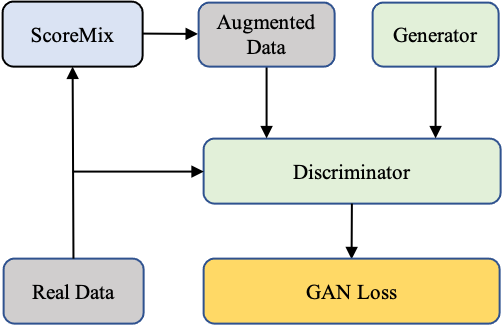}
         \caption{}
         \label{fig:p1}
     \end{subfigure}
     \hspace{55pt}    
     \begin{subfigure}[b]{0.35\textwidth}
         \centering
         \includegraphics[width=\textwidth]{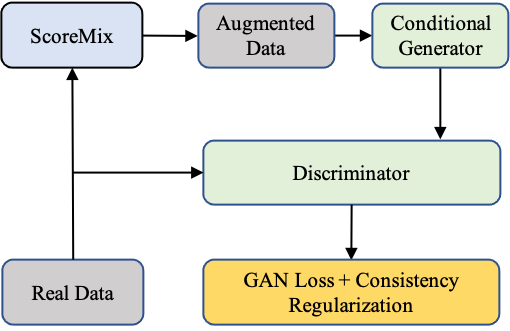}
         \caption{}
         \label{fig:p2}
     \end{subfigure}
\caption{
Overall pipeline of training Generative Adversarial Networks with the proposed ScoreMix method.
The noise input to the generator is omitted for simplicity. 
We synthesize augmented samples with a score-based mixing function (Section~\ref{sec:32}).
(a) For unconditional image synthesis, we provide the discriminator with the augmented samples as ground truth.
(b) For image-to-image translation, we feed the original samples and augmented samples to the generator.
}
\label{fig:pipeline}
\end{figure*}

\section{Proposed Method}

The remainder of this section is organized as follows. 
Section~\ref{sec:31} presents the overall data augmentation pipeline. 
In Section~\ref{sec:32}, we introduce the proposed score-based augmentation function.
In Section~\ref{sec:33}, we describe how to train the score estimation network.
Section~\ref{sec:34} provides a comprehensive comparison between our approach and existing mixing-based methods.

\subsection{Overall Augmentation Pipeline}
\label{sec:31}

GANs~\cite{goodfellow2014generative} aim to model a data distribution $\pdata$ with a generator and a discriminator. 
GAN training is a two-player game: the generator learns to generate samples that resemble the real samples, whereas the discriminator learns to discriminate between the real and fake samples. 
Formally, the objective function of GANs can be formulated as:
\begin{equation}
\label{eq:gan}
\min_{G} \max_{D} \E_{\x} [\log D(\x)] + \E_{\z} [\log(1 - D(G(\z)))],
\end{equation}
where $\x \in  \mathbf{R}^{n}$ is drawn from $\pdata$, and $z$ is drawn from a prior distribution. We use $\G$ and $\D$ to denote the generator and discriminator, respectively. 
Given a small training dataset, the discriminator $\D$ is prone to overfit the real samples. 
If this happens, the GAN training will collapse quickly since $\D$ does not provide useful supervision. 

To address this issue, we propose a mixing-based data augmentation strategy.
For unconditional GANs, as illustrated in Figure~\ref{fig:p1}, we create augmented samples using the combinations of real samples.
$\D$ regards the augmented samples as the ground truth data during training.
When $\G$ conditions on the real sample, as shown in Figure~\ref{fig:p2}, we augment the input data for $\G$.
We use this strategy for practical applications such as image-to-image translation tasks.

The core component of the augmentation strategy is the mixing function. 
It increases the diversity of training data and prevents overfitting. 
Recall that one of the most classical mixing-based augmentation strategies is the mixup method~\cite{zhang2017mixup}, which uses the linear interpolation of samples.
Concretely, given two real samples $\xa, \xb \sim \pdata$, its mixing function can be formulated as:
\begin{equation}
\label{eq:mixup}
\xmixup = \mixup(\xa, \xb) = \weight \cdot \xa + (1 - \weight) \cdot \xb,
\end{equation}
where $\weight \in [0,1]$ is the interpolation weight sampled from the Beta distribution, \ie, $\weight \sim \text{Beta}(\alpha, \alpha)$ with $\alpha \in (0,  +\infty)$ being a hyperparameter. 
As mentioned in Section~\ref{sec:21}, the mixup method may introduce characteristic image artifacts, leading to manifold intrusion.
In the following, we introduce the proposed mixing function to address this issue.

\subsection{ScoreMix Augmentation}
\label{sec:32}
We first introduce the notion of the Stein score.
According to Stein’s method~\cite{stein_bound_1972}, given a smooth data distribution $\pdata$ supported on the set of real numbers, we define the Stein score of a sample $\x \sim \pdata$ as follows:
\begin{equation}
\label{eq:stein_score}
	\stein_{\pdata}(\x) := \nabla_{\x} \pdata(\x) / p(\x) = \nabla_{\x} \log \pdata(\x).
\end{equation}

Equation~\ref{eq:stein_score} shows that the Stein score can be interpreted as the gradients of the logarithmic distribution function with respect to data. 
Here we assume that a deep neural network $\snet$ can estimate the Stein score (discussed later in Section~\ref{sec:33}). 
Given two real samples $\xa$ and $\xb$, we compute the initial value $\xmixup$ using the mixup method~\cite{zhang2017mixup} by Equation~\ref{eq:mixup}.
Then, we search for $\xremix$ with the following objective:
\begin{equation}
\label{eq:remix}
	\xremix = \argmin{\textcolor{black}{\x \in U(\xmixup)}}~\ltwo{\snet(\textcolor{black}{\x})} := \scoremix(\xmixup; \snet),
\end{equation}
where $\ltwo{\cdot}$ denotes the Euclidean L2 norm.
We use the gradient descent method to solve Equation~\ref{eq:remix}. 
The solution $\xremix$ is a stationary point of the logarithmic distribution function $\log \pdata(\x)$. 
This implies $\xremix$ is also a stationary point of the distribution function $\pdata(\x)$. 
Since $\xmixup$ produced from the mixup method is close to the manifold of $\pdata$~\cite{zhang_how_2021}, $\xremix$ is a local maximum almost surely. 
Hence, the augmented sample~$\xremix$ has the maximum likelihood that belongs to $\pdata$ in a local neighborhood.

\textcolor{black}{
In Equation~\ref{eq:remix}, the initial value $\xmixup$ provides a strong image prior, which prevents the score-based model from producing samples that deviate from the real samples.
This is particularly useful when the training data is insufficient.
Moreover, using $\xmixup$ as initialization accelerates the augmentation process.
This allows the score estimation network to optimize an augmented sample from an intermediate result, which is significantly faster than the initialization with random noise in the existing methods (\eg, \cite{song_generative_2019,song_improved_2020,song_score-based_2021}).
}

\textcolor{black}{
The proposed ScoreMix augmentation follows the Vicinal Risk Minimization principle~\cite{chapelle2000vicinal}.
Similar to the mixup method~\cite{zhang2017mixup}, given a training data set $\mathcal{D}$, we optimize the empirical vicinal risk.
Concretely, we define the vicinity distribution $\nu$ as follows:
\begin{align}
\nu(\xremix | \xii)
&\!=\!\frac{1}{n} \sum_j^n \underset{\lambda}{\mathbb{E}}\left[\delta\left(\xremix\! = \!\scoremix( \lambda \cdot \xii + (1-\lambda) \cdot \xjj ; S_{\theta})\right)\right] \nonumber \\
&\!=\!\frac{1}{n} \sum_j^n \underset{\lambda}{\mathbb{E}}\left[\delta\left(\xremix\! = \!\scoremix( \xmixup ; \snet )\right)\right],
\end{align}
where $\delta(\xremix \! = \! \x)$ is a Dirac mass centered at $\x$.
Compared with the mixup method, our approach produces augmented samples closer to the training data distribution.
This is because $\scoremix$ produces a local maximum of the distribution, which implies that  $\pdata(\xremix) \geq \pdata(\xmixup)$.
}

Algorithm~\ref{alg:scoremix} shows the main steps to augment training data with the proposed method.
We use Equation~\ref{eq:mixup} to compute initial values and then use Equation~\ref{eq:remix} to push the initialized augmented samples towards the real data manifold.
The proposed augmentation method can be plugged into existing GAN-based methods without changing the underlying algorithms.

{
\linespread{1.2} 
\begin{algorithm}[t]
\SetAlgoLined
 \KwIn{\,Image samples $\xa$ and $\xb$ \; ~~~~~~~~~~~Score estimation network $\snet$\;
 ~~~~~~~~~~~Learning rate $\eta$, optimization step $N$\;
 ~~~~~~~~~~~Distribution $\text{Beta}(\alpha, \alpha)$.}
 \KwOut{Augmented sample $\xremix$.}
 Sample interpolation weight $\weight \sim \text{Beta}(\alpha, \alpha)$\;
 Compute the mixed sample $\xmixup = \weight \cdot \xa + (1 - \weight) \cdot \xb$\;
 Initialize $\x \leftarrow \xmixup$, and $i \leftarrow 0$\;
 \While{$i < N$}{
 Calculate loss $\mathcal{L}(\xremix) = \ltwo{\snet(\xremix)}$\;
 Calculate gradient $\nabla_{\xremix}\mathcal{L}(\xremix)$\;
 $\xremix \leftarrow \xremix - \eta \cdot \nabla_{\xremix}\mathcal{L}(\xremix)$\;
 $i \leftarrow i+1$\;
 }
 return $\xremix$

 \caption{The ScoreMix augmentation method.}
 \label{alg:scoremix}
\end{algorithm}
}

\subsection{Score Estimation Network}
\label{sec:33}

We build a score estimation network $\snet$ to estimate the Stein score $\stein_{\pdata}(\x) $.
Following the idea of score matching~\cite{hyvarinen2005estimation}, we do not estimate $\pdata(\x)$ first.
Instead, we directly optimize the network parameters $\theta$ by minimizing:
\begin{equation}
\label{eq:sm}
\E_{\x} [\| S_{\theta}(\x) - \nabla_{\x} \log \pdata (\x) \|_{2}^{2}],
\end{equation}
where we have $\x \in  \mathbf{R}^{n}$ drawn from $\pdata$.
However, in Equation~\ref{eq:sm}, the term $\nabla_{\x} \log \pdata(\x)$ is computationally intractable for real-world data distributions.
\textcolor{black}{To address this problem, we use the denoising score matching (DSM) method~\cite{vincent_connection_2011}, which is robust in low-density data regions.
The DSM method perturbs training samples with a prior distribution whose supports span the whole image space $\mathbf{R}^{n}$.
This strategy enforces that the perturbed training data distribution has full support over $\mathbf{R}^{n}$.
Hence, the score-based model gets stable training signals~\cite{hyvarinen2005estimation} even in low-density data regions.}

Concretely, we perturb $\x$ with Gaussian noise~$\mathcal{N}(0, \var)$ which has a zero mean and variance $\var$. 
Then, we use $\snet$ to estimate the Stein score of the perturbed sample $\xt$.
Let $\ppdata$ denote the perturbed data distribution, and the objective can be formulated:
\begin{equation}
\E_{\xt} [\| S_{\boldsymbol{\theta}}(\xt) - \nabla_{\xt} \log q_{\sigma}(\xt \mid \x) \|_{2}^{2}],
\end{equation}
where we have $\xt \sim \ppdata(\xt \ | \ \x)=\mathcal{N}(\x, \sigma^{2})$.
The objective is tractable since we have $\nabla_{\xt} \log q_{\sigma}(\xt \mid \x) = - \frac{1}{\var} \cdot (\xt - \x)$.
This immediately implies that we can rewrite the objective as:
\begin{equation}
\label{eq:dsm_gaussian}
\E_{\xt} [\| S_{\boldsymbol{\theta}}(\xt) + \frac{\xt - \x}{\var} \|_{2}^{2}].
\end{equation}

We could minimize Equation~\ref{eq:dsm_gaussian} to train the estimation network.
The optimal solution $\sneto(\x)$ converges to $\nabla_{\x} \log \pdata (\x)$  as long as the noise variance $\var$ is small enough \cite{vincent_connection_2011}.
To train the estimation network $\snet$ more efficiently, we propose to perturb the data samples with multi-scale Gaussian noises.
Specifically, the variances of the multi-scale noises are a geometric sequence $\{\sigma_i\}_{i=1}^N$, and we have:
\begin{equation}
	\sigma_n = \sigma_1 \cdot \gamma^{n - 1},
\end{equation}
where $\gamma$ denotes the common ratio, and $n \in [0, N]$. 
Without loss of generality, we assume that $0 < \gamma < 1$.
Then, $\sigma_1$ has the maximal value, and  $\sigma_N$ has the minimal value.
On the one hand,  the perturbed sample with the minimal noise scale is close enough to the real sample.
We can regard $\nabla_{\xt} \log q_{\sigma_{N}}(\xt \mid \x)$ as the Stein score of $\x$ with negligible estimation error.
On the other hand, the Stein score of the perturbed samples with a larger noise scale is easier to estimate.
Hence, these samples help to stabilize the training process.
Concretely, we train the estimation network $\snet$ by minimizing:
\begin{align}
\label{eq:dsm}
\loss{\textrm{SCORE}} &= \sum_{i=1}^{N} \omega_i \cdot \E_{\xt}\ [\| S_{\boldsymbol{\theta}}(\xt, \sigma_i) -  \nabla_{\xt} \log q_{\sigma}(\xt \mid \x) \|_{2}^{2}] \nonumber \\
&= \sum_{i=1}^{N} \sigma_i^2 \cdot \E_{\xt}\ [\| S_{\boldsymbol{\theta}}(\xt, \sigma_i) + \frac{\xt - \x}{\vari} \|_{2}^{2}],
\end{align}
where $\xt \sim q_{\sigma_i}(\xt \ | \ \x)=\mathcal{N}(\x, \sigma_i^{2})$.
The score estimation network $\snet$ conditions on the noise variation $\sigma_i^{2}$ during training.
Note that in Algorithm~\ref{alg:scoremix}, we only sample the noises with the minimal variation $\sigma_N^{2}$.
The weighting term $\omega_i$ keeps the expectation values in Equation~\ref{eq:dsm} in the same order of magnitude, which is beneficial for model training~\cite{song_improved_2020}.

We scale the pixel values of samples to the range $[0, 1]$ and set $\sigma_N=0.01$.
We set the value of $\sigma_1$ to the maximum Euclidean distance between all pairs of training samples.
One may determine the value of the common ratio $\gamma$ using the heuristic rules proposed by Song \etal~\cite{song_improved_2020}.
However, we simply use $\gamma=0.99$ and find this setting works well in the following experiments.

\subsection{Comparison with Existing Methods}
\label{sec:34}

\begin{figure}[t]
\begin{center}
\includegraphics[width=1.0\linewidth]{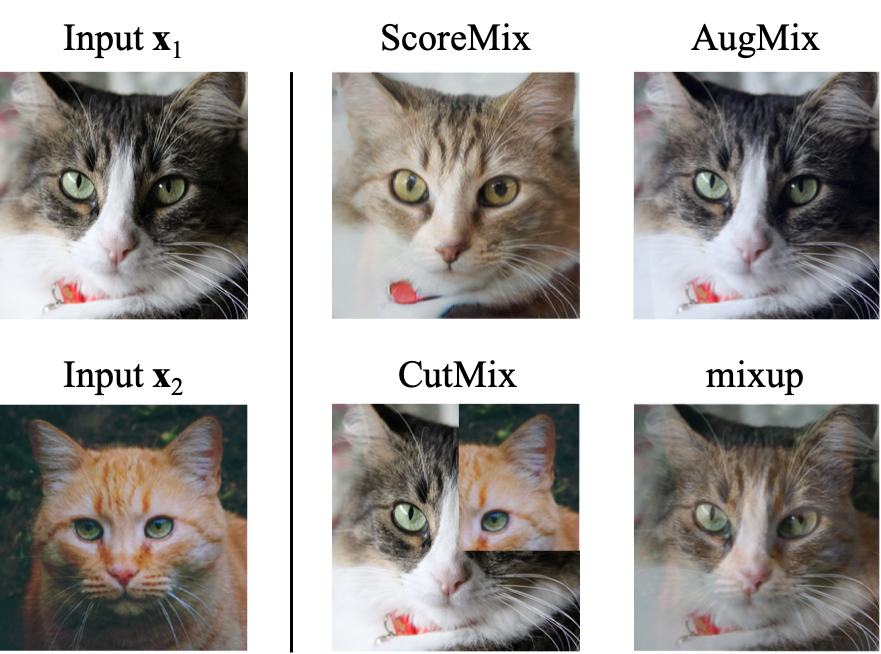}
\end{center}
   \caption{
Visual comparison of the augmented images. 
Mixup~\cite{zhang2017mixup}, CutMix~\cite{yun_cutmix_2019}, and ScoreMix (ours) sythesize the augmented images using the combinations of inputs $\x_{1}$ and $\x_{2}$. 
AugMix~\cite{hendrycks_augmix_2020} generates the augmented image with input $\x_{1}$. 
The input images are from the AFHQ v2 dataset~\cite{choi2020stargan}. }
\label{fig:mix}
\end{figure}

In this subsection, we show the advantages of the ScoreMix method by making comparisons with previous mixing-based methods:

\begin{itemize}
	\item 
	The proposed ScoreMix method does not require any data-specific hyperparameter tuning. 
	In contrast, previous methods such as AugMix~\cite{hendrycks_augmix_2020} build data augmentation pipelines consisting of numerous hyperparameters.
	These hyperparameters need to be tuned for different tasks. 
	Otherwise, the augmentation methods may even hurt model performance.
	\item 
	We do not rely on any pre-trained model during training. 
	Instead, we first train a score estimation network and then use it to guide the augmentation process.
	Methods like SaliencyMix~\cite{uddin_saliencymix_2021} and Puzzle Mix~\cite{kim_puzzle_2020} generate augmented data with the guidance of a pre-trained saliency detection model. 
	But how to obtain an accurate saliency detection model in the data-limited regime remains a challenging problem.
	The requirement of accurate pre-trained models hinders the application of these methods in real-world scenarios.
	\item 
	We do not require any human annotations for data augmentation.
	That means the ScoreMix method is readily applicable to unsupervised image synthesis tasks.
	In contrast, the SuperMix method~\cite{dabouei_supermix_2021} and hardmix method~\cite{kalantidis_hard_2020} improve the data augmentation strategies with label information, which may be expensive or difficult to obtain in practice.
\end{itemize}

Figure~\ref{fig:mix} shows some examples of the augmented results from our approach and several existing methods~\cite{zhang2017mixup,yun_cutmix_2019,hendrycks_augmix_2020}.
The example from the mixup method~\cite{zhang2017mixup} exhibits characteristic artifacts.  
Two cats in the augmented image are displaced in relation to each other. 
For the CutMix method~\cite{yun_cutmix_2019}, it is easy to notice the edges of the pasted patch, and the augmented image looks unrealistic. 
We note that the mixup and CutMix methods augment samples using image arithmetic operations (e.g., addition and multiplication), which inevitably produces unrealistic details. 
The AugMix method~\cite{hendrycks_augmix_2020} aims to generate augmented images without deviating too much from the original ones. 
However, the augmented result seems to be too similar to the original image.
By contrast, the example from our approach is distinguished from the original image. 
We find that the fur color and pupil shape of the cat are changed. 
Our approach also subtly changes the shape of the stripes on the forehead of the cat. 
Moreover, our result expresses natural textures and details.

\section{Experiments and Analysis}

\subsection{Implementation Details}
To generate the augmented sample $\xremix$, we use the Adam optimizer with a learning rate of 0.005 and linear decay scheduling.
We use 100 gradient descent steps for the optimization and then finetune $\xremix$ by score-based denoising.
The score estimation network optimizes $\xremix$ and approximates the denoising process at $t_{0} = 0.25$.
Similar to the mixup method~\cite{zhang2017mixup}, we set the hyperparameter $\alpha=0.2$ for the beta distribution.
We use these settings for all the experiments without hyperparameter tuning.

We build the score estimation network following the architecture design in \cite{song_generative_2019}.
Specifically, we adopt the RefineNet~\cite{lin2017refinenet} architecture and use the conditional instance normalization~\cite{dumoulin2016learned} for every convolutional layer. 
To train the score estimation network, we use the Adam optimizer with a learning rate of 1e-4. 
The training batch size is set to 32.

\subsection{Datasets}

In this section, we introduce the datasets used in the following experiments.

The Flickr-Faces-HQ (FFHQ) dataset~\cite{karras_style-based_2019} is a benchmark dataset for unconditional image synthesis. 
This dataset consists of 70~thousand high-quality human face images with considerable variations. 
The images are numbered from 0 to 69,999, and we use the first one thousand images for model training.

The ImageNet dataset~\cite{krizhevsky_imagenet_2012} consists of various images organized into 1,000 classes.
It is a standard benchmark for conditional image synthesis.
In this work, we train the generation models with 10\% training images per class.

The MetFace dataset~\cite{karras_training_2020} consists of human face images from the Metropolitan Museum of Art.
There are 1,336 manually selected images which depict artworks such as paintings, drawings, and statues.
In the following, we train the models with the first one thousand images.

The Animal Faces-HQ v2 (AFHQ) dataset~\cite{choi2020stargan} provides animal faces of three domains: cat, dog, and wildlife. 
Each category contains about 5,000 images. 
Choi \etal~\cite{choi2020stargan} split this dataset into 14,336 samples for training and 1,467 samples for testing.
We randomly choose 500 training images for each class, which is about 10\% of the full training set.

\subsection{Unconditional Image Synthesis}

We first verify the proposed ScoreMix method by unconditional image synthesis.
In this task, we train GANs to map random noise to image samples.
We use the StyleGAN2~\cite{karras2020analyzing} as the baseline network, and the output resolution is $256 \times 256$.
We implement the baseline using the released source codes\footnote{\url{https://github.com/rosinality/stylegan2-pytorch}}.

We evaluate the quality of synthesized images using Fr\'echet Inception Distance (FID)~\cite{heusel2017gans} and Kernel Inception Distance (KID)~\cite{binkowski_demystifying_2018}.
The FID metric measures the Wasserstein distance between two image sets. 
We use the FID score between the real images and the synthesized images to evaluate the sample quality: a lower score indicates better results.
The KID metric measures the squared maximum mean discrepancy between the representations of the real and synthesized image sets.
Similarly, a lower KID indicates that the results are better.

In the experiments, we find that the GAN training is unstable when the training data is limited. 
The FID score starts to deteriorate after a certain number of training iterations. 
That means the overfitting problem occurs. Then, the generated samples become meaningless noises. 
Such observations are also reported in previous methods~\cite{brock_large_2019,karras_training_2020,zhao2020differentiable}.
Hence, we perform the early stopping strategy to train all the models. 
Specially, we compute the FID score every 100, 000 iterations during training. 
The training is stopped when the FID score increases in five consecutive epochs.
Then, we use the training snapshot with the best FID score for evaluation.

\begin{figure}[t!]
\begin{center}
    \includegraphics[width=\linewidth]{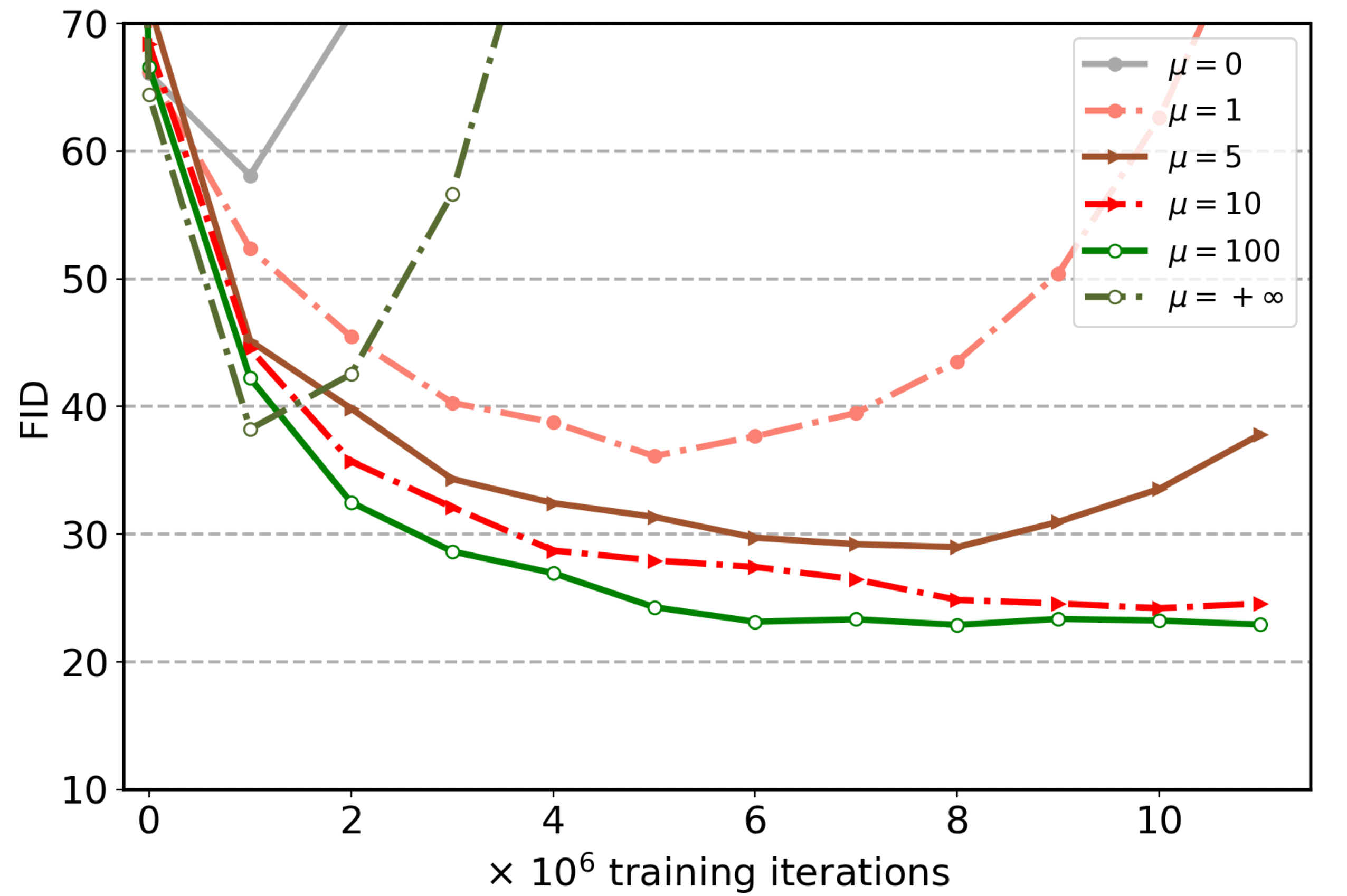}
\end{center}
\caption{
Ablation of the augmentation ratio $\mu$.
We train the StyleGAN2 model~\cite{karras2020analyzing} with the proposed ScoreMix method on the FFHQ dataset~\cite{karras_style-based_2019} under 1K data setting.
We compute Fr\'echet Inception Distance~\cite{heusel2017gans} (FID, lower is better) during training.}
\label{fig:ratio}
\end{figure}

{
\renewcommand{\arraystretch}{1.1}
\begin{table}[t!]
\begin{center}
\caption{
Ablation of baseline method. We train different baselines on the FFHQ dataset~\cite{karras2020analyzing} under 1K data setting. We report FID~\cite{heusel2017gans} (lower is better). The former/later number in the table cell denotes the FID score of the baseline method without/with the proposed scheme.
}
\label{tab:baseline}%
    \begin{tabular}{clcc}
    \toprule
    &~~~~~Network & \small{Preprocessing method} & ~~FID$\downarrow$~~\\
    \midrule
    ($\textrm{a}$) & StyleGAN2~\cite{karras2020analyzing} &  none & 58.11\,/\,24.21 \\
    ($\textrm{b}$) & $\text{StyleGAN2}_{\textrm{narrower}}$ &  none &  55.91\,/\,25.04 \\
    ($\textrm{c}$) & $\text{StyleGAN2}_{\textrm{wider}}$ &  none &  64.38\,/\,23.76 \\
    \midrule
    ($\textrm{d}$) & StyleGAN2~\cite{karras2020analyzing} &  DiffAug~\cite{zhao2020differentiable} & 26.63\,/\,21.13 \\
    ($\textrm{e}$) & $\text{StyleGAN2}_{\textrm{narrower}}$ &  DiffAug~\cite{zhao2020differentiable} &  29.62\,/\,21.26 \\
    ($\textrm{f}$) & $\text{StyleGAN2}_{\textrm{wider}}$ &  DiffAug~\cite{zhao2020differentiable} &  26.13\,/\,19.41 \\
    \bottomrule
    \end{tabular}%
\end{center}
\end{table}%
}

{
\renewcommand{\arraystretch}{1.25}
\setlength\tabcolsep{3pt}
\begin{table}[t!]
\begin{center}
\caption{
Unconditional image synthesis results of different methods on the FFHQ~\cite{karras2020analyzing} and MetFace~\cite{karras_training_2020} datasets under 1K data setting. We report FID~\cite{heusel2017gans} (lower is better) and KID~\cite{binkowski_demystifying_2018} (lower is better). ScoreMix is our proposed method. $\text{ScoreMix}_{\text{LC}}$ and $\text{ScoreMix}_{\text{ADA}}$ combine it with the regularization methods in \cite{tseng_regularizing_2021} and \cite{karras_training_2020}, respectively.
}
\label{tab:com1}%
\smallskip\noindent
\resizebox{\linewidth}{!}{

    \begin{tabular}{lccccc}
    \toprule
    \multicolumn{1}{c}{\multirow{2}[4]{*}{Method}} & \multicolumn{2}{c}{FFHQ-1K} &       & \multicolumn{2}{c}{MetFace-1K} \\
\cmidrule{2-3}\cmidrule{5-6}          & \footnotesize{FID\,$\downarrow$}   & \multicolumn{1}{l}{\footnotesize{KID$\times 10^{3}\downarrow$}} &       & \multicolumn{1}{l}{\footnotesize{FID\,$\downarrow$}} & \multicolumn{1}{l}{\footnotesize{KID$\times 10^{3}\downarrow$}} \\
    \midrule
	{\fontsize{7.5pt}{10pt}\selectfont GAN: \makebox[17mm][l]{StyleGAN2}~\cite{karras2020analyzing}} & 58.11 & 40.32 &     &  64.58     &  40.60 \\
    {\fontsize{7.5pt}{10pt}\selectfont \makebox[24mm][l]{\textcolor{black}{score-based: NCSNv2}}~\cite{song_improved_2020}} &  38.20 & 21.73  &       & 44.56   & 19.48  \\
    {\fontsize{7.5pt}{10pt}\selectfont \makebox[24mm][l]{\textcolor{black}{score-based: SMLD}}~\cite{song_score-based_2021}} & 31.23 & 17.19  &       & 35.70   & 13.71  \\
    \midrule
    {\fontsize{7.5pt}{10pt}\selectfont Baseline: ~~~\cite{karras2020analyzing}~~+~~\cite{zhao2020differentiable}} & 26.13 & 14.56 &     &  34.60     &  13.87 \\
    {\fontsize{7.5pt}{10pt}\selectfont \makebox[21mm][l]{Baseline + mixup}~\cite{zhang2017mixup}} & 25.25 & 13.87 &       &   31.93    &   11.81\\
    {\fontsize{7.5pt}{10pt}\selectfont \makebox[21mm][l]{Baseline + CutMix}~\cite{yun_cutmix_2019}} & 24.72 & 13.37 &       &  30.11     &  12.27 \\
    {\fontsize{7.5pt}{10pt}\selectfont \makebox[21mm][l]{Baseline + AugMix}~\cite{hendrycks_augmix_2020}} & 22.95 & 9.46  &       &  25.91     &  6.53 \\
    \midrule
    {\fontsize{7.5pt}{10pt}\selectfont Baseline + ScoreMix} & \textbf{19.41} & \textbf{5.65}  &     &  \textbf{22.13}   & \textbf{4.51}  \\
    {\fontsize{7.5pt}{10pt}\selectfont Baseline + \textcolor{black}{$\text{ScoreMix}_{\text{ADA}}$}} & 18.74 & 4.38  &       &  \textbf{19.12}  & \textbf{2.97}  \\
    {\fontsize{7.5pt}{10pt}\selectfont Baseline + \textcolor{black}{$\text{ScoreMix}_{\text{LC}}$}} & \textbf{18.29} & \textbf{3.91} &       &  19.42   & 3.35  \\
    \bottomrule
    \end{tabular}

}
\end{center}
\end{table}
}

\vspace{2mm}
\noindent \textbf{Ablation: augmentation ratio.}
We aim to determine the optimal amount of augmented data. 
Intuitively, this amount depends on the amount of real data. 
Hence, we study the effect of the augmentation ratio $\mu$, \ie, the ratio of augmented to real data. 
In this ablation study, we investigate the following two cases:

\begin{itemize}
	\item 
	The augmentation ratio $\mu$ is static.
	We produce the augmented data before the training starts.
	The amount of training data increases by a factor of ($\mu + 1$). 
	\item 
	The augmentation ratio $\mu$ grows during training. 
	We augment the training batch with a certain probability at each iteration. 
	The amount of augmented data will increase as the training goes on. 
	Considering that the number of training iterations is huge, we assume the augmentation ratio approaches infinity.
\end{itemize}

The results are shown in Figure~\ref{fig:ratio}. 
Overall, GANs benefit from augmented training data. 
We observe that the FID scores improve, and most of the training processes become more stable.
These results show that augmented data can improve model performance.
Models with a growing augmentation ratio still collapse quickly.
One possible reason is that the amount of augmented data is overwhelming.
The models have to neglect the real data to minimize the training loss.
This phenomenon indicates that too many augmented samples is not helpful to model training.

The model with a larger static augmentation ratio renders better results, but we observe diminishing returns in terms of sample quality.
When we crease augmentation ratio $\mu$ from 0 to 10, the FID score decreases from 58.10 to 24.21.
Whereas, increasing $\mu$ to 100 only makes the FID score 1.31 lower.
To balance the trade-off between sample quality and computational cost, we set $\mu=10$ for the following tasks.

\vspace{2mm}
\noindent \textbf{Ablation: baseline model.} 
In this ablation study, we investigate whether the proposed method consistently improves model performance across different baseline models or not. 
Specifically, we consider the following two types of variants:

\begin{itemize}
	\item 
	Baselines with different model capacity. 
	We change the network architecture to increase/reduce model capacity. 
	Concretely, we double/halve the network width (the number of feature channels). 
	\item 
	Baselines with additional augmentation method. 
	We add the DiffAug method~\cite{zhao2020differentiable}, which consists of a family of the traditional augmentation operations~\cite{lecun_gradient-based_1998,bengio_deep_2011,krizhevsky_imagenet_2012}, to the augmentation pipeline.
\end{itemize}

The evaluation results of the variants are shown in Table~\ref{tab:baseline}.
We use the subscript ${\textrm{wider}}$/${\textrm{narrower}}$ to indicate more/fewer feature channels.
For all the variants of baselines, the proposed ScoreMix method significantly improves model performance. 
In addition, combining our approach with the DiffAug method~\cite{zhao2020differentiable} yields further performance gain.
$\text{StyleGAN2}_{\textrm{wider}}$ performs the best among different network architectures.
This shows that increasing model capacity leads to better results if we augment the training data properly.
The variant (e), \ie, wider StyleGAN model + DiffAug + ScoreMix, achieves the best FID score. 
In the following, we consider this variant as the baseline method for unconditional image synthesis unless otherwise specified.

{
\renewcommand{\arraystretch}{1.15}
\captionsetup{labelfont={color=black},font={color=black}}
\setlength\tabcolsep{2.5pt}
\begin{table}[t!]
\centering
\caption{
FID~\cite{heusel2017gans} scores (lower is better) as a function of training set size on the FFHQ~\cite{karras2020analyzing} dataset. ScoreMix is our proposed method. $\text{ScoreMix}_{\text{LC}}$ and $\text{ScoreMix}_{\text{ADA}}$ combine it with the regularization methods in \cite{tseng_regularizing_2021} and \cite{karras_training_2020}, respectively.}
\label{tab:2.3.2.1}%
\smallskip\noindent
\resizebox{\linewidth}{!}{
    
    \begin{tabular}{lccccccccccc}
    \toprule
    \multicolumn{1}{c}{Method} & 70k   &       & 35k   &       & 10k   &       & 5k    &       & 2k    &       & 1k \\
    \midrule
    {\fontsize{7.5pt}{10pt}\selectfont Baseline:~~\cite{karras2020analyzing}~+~\cite{zhao2020differentiable}} & 4.56  &       & 5.85  &       & 8.85  &       & 13.44 &       & 17.93 &       & 26.13 \\
    \midrule
    {\fontsize{7.5pt}{10pt}\selectfont Baseline + ScoreMix} & 4.32  &       & 5.36  &       & 7.57  &       & 10.46 &       & 13.45 &       & 19.41 \\
    {\fontsize{7.5pt}{10pt}\selectfont Baseline + $\text{ScoreMix}_{\text{ADA}}$} & 4.25  &       & 5.19  &       & 7.08  &       & 9.84  &       & 12.61 &       & 18.74 \\
    {\fontsize{7.5pt}{10pt}\selectfont Baseline + $\text{ScoreMix}_{\text{LC}}$} & 4.29  &       & 5.03  &       & 6.85  &       & 9.54  &       & 12.04 &       & 18.29 \\
    \bottomrule
    \end{tabular}%

}
\end{table}
}

\vspace{2mm}
\noindent \textbf{Evaluation results.} 
We evaluate the proposed ScoreMix method against existing mixing-based approaches, including the mixup~\cite{zhang2017mixup}, CutMix~\cite{yun_cutmix_2019}, and AugMix~\cite{hendrycks_augmix_2020} methods.
Note that for the mixup and CutMix methods, we use them to augment both the real and fake samples for the discriminator.
This strategy is necessary for them to avoid the leaking problem~\cite{karras_training_2020}.

Table~\ref{tab:com1} reports the FID and KID scores evaluated on the FFHQ~\cite{karras2020analyzing} and MetFace~\cite{karras_training_2020} datasets.
\textcolor{black}{
The score-based models achieve better FID/KID scores than the GAN model~\cite{karras2020analyzing} under the 1K data setting.
This provides empirical evidence that the score-based methods are more robust in data-limited regimes.
Our approach, combining the score-based data augmentation with GANs, performs best in terms of FID and KID on both datasets.
}

\textcolor{black}{
In addition, $\text{ScoreMix}_{\text{LC}}$ combines the proposed method with the LC regularization in our prior work~\cite{tseng_regularizing_2021}, and $\text{ScoreMix}_{\text{ADA}}$ combines the proposed method with the adaptive discriminator augmentation (ADA) method~\cite{karras_training_2020}.
The comparison results in Table~\ref{tab:com1} show that appending the regularization method in \cite{tseng_regularizing_2021} or \cite{karras_training_2020} to ScoreMix both leads to further improvements in quantitative metrics.}

\begin{figure*}[t!]
\begin{center}
\includegraphics[width=1.0\linewidth, height=455pt]{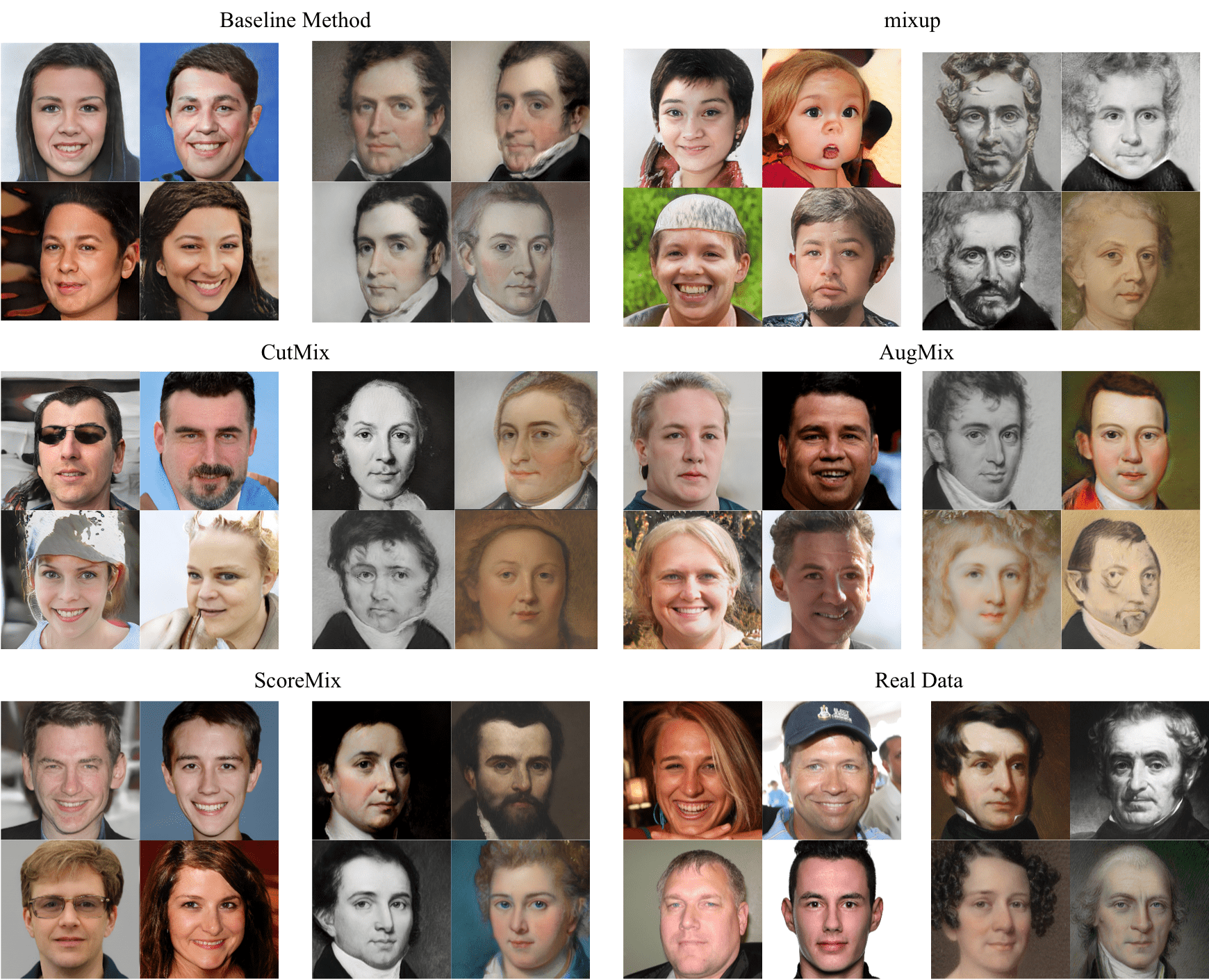}
\end{center}
\caption{
Unconditional FFHQ~\cite{karras2020analyzing}  (photographs) and MetFace~\cite{karras_training_2020} (portrait painting) samples generated by different methods under 1K data setting. We use the StyleGAN2 model~\cite{karras2020analyzing} with the DiffAug scheme~\cite{zhao2020differentiable} as the baseline. The results are synthesized by the baseline, baseline + mixup~\cite{zhang2017mixup}, baseline + CutMix~\cite{yun_cutmix_2019}, baseline + AugMix~\cite{hendrycks_augmix_2020}, and baseline + ScoreMix (ours).
}
\label{fig:vis1}
\end{figure*}

We visualize some samples synthesized by different methods in Figure~\ref{fig:vis1}.
Our synthesized results are diverse and contain fewer artifacts.
This indicates the generated distribution of our results is similar to the real distribution.

\textcolor{black}{
Table~\ref{tab:2.3.2.1} shows the FID~\cite{heusel2017gans} scores with different amounts of training data (from 70K to 1K) on the FFHQ dataset~\cite{karras2020analyzing}.
The results show that our approaches consistently improve the baseline method under all the data settings.
As the number of training samples increases, the baseline method benefits less from our approach.
However, under the 70K setting, ScoreMix still improves the FID score by 5.3\% (4.56 $\rightarrow$ 4.32). 
}

\vspace{2mm}
\noindent \textcolor{black}{\textbf{Computational cost.} 
The proposed ScoreMix method has a reasonable computational cost.
In our experiments, training the baseline method (StyleGAN2~\cite{karras2020analyzing} + DiffAug~\cite{zhao2020differentiable}) takes 69.3 hours on 4 Tesla V100S GPUs.
We follow the setting in \cite{karras_training_2020} that stops the GAN training when 25,000k real samples are shown to the discriminator.
The training that combines the baseline method with our approach takes 76.0 hours, which increases the wall-clock time by around 9.5\%.
In addition, we take about 120 hours to train the score estimation network. 
During inference, our approach requires no additional computational cost.
}

\textcolor{black}{\subsection{Class-Conditional Image Synthesis}}

\textcolor{black}{
We also evaluate the proposed method on class-conditional image synthesis using the ImageNet dataset~\cite{russakovsky_imagenet_2015}, where we train GANs to produce samples with a specified class based on a given label.
Since labeled data is usually expensive to collect, solving this conditional task with limited data is important for practical applications.
We use BigGAN~\cite{brock_large_2019} with the DiffAug method~\cite{zhao2020differentiable} as the baseline.
The output resolution is $128 \times 128$.
Note that we implement the baseline method on GPU, and we reduce the training batch size to 512, which is 4 times less than the size in the original work~\cite{brock_large_2019}.
}

{
\renewcommand{\arraystretch}{1.25}
\captionsetup{labelfont={color=black},font={color=black}}
\begin{table*}[t!]
\centering
\caption{
Class-conditional image synthesis results of different methods on the ImageNet dataset~\cite{russakovsky_imagenet_2015}. We report FID~\cite{heusel2017gans} (lower is better) and IS~\cite{binkowski_demystifying_2018} (higher is better). We implement BigGAN~\cite{brock_large_2019} on GPU with a training batch size of 512. ScoreMix is our proposed method. $\text{ScoreMix}_{\text{LC}}$ and $\text{ScoreMix}_{\text{ADA}}$ combine it with the regularization methods in \cite{tseng_regularizing_2021} and \cite{karras_training_2020}, respectively.
}
\label{tab:2.3.3.1}%
\resizebox{0.9\linewidth}{!}{

	\begin{tabular}{lccccccccccc}
    \toprule
    \multicolumn{1}{c}{\multirow{2}[4]{*}{Methods}} & \multicolumn{2}{c}{100\% data} &       & \multicolumn{2}{c}{50\% data} &       & \multicolumn{2}{c}{25\% data} &       & \multicolumn{2}{c}{10\% data} \\
\cmidrule{2-3}\cmidrule{5-6}\cmidrule{8-9}\cmidrule{11-12}          & FID\,$\downarrow$   & IS\,$\uparrow$    &       & FID\,$\downarrow$   & IS\,$\uparrow$    &       & FID\,$\downarrow$   & IS\,$\uparrow$    &       & FID\,$\downarrow$   & IS\,$\uparrow$ \\
    \midrule
    Baseline:~~\cite{brock_large_2019}\,\,\,\,+\,\,\cite{zhao2020differentiable} & 9.08  & 91.82  &       & 12.88  & 83.27  &       & 17.10  & 62.15  &       & 49.67  & 21.30  \\
    \midrule
    \makebox[22.5mm][l]{Baseline + mixup}~\cite{zhang2017mixup} & 9.01  & 91.19  &       & 12.70  & 84.05  &       & 16.63  & 64.30  &       & 42.19  & 25.41  \\
    \makebox[22.5mm][l]{Baseline + CutMix}~\cite{yun_cutmix_2019} & 9.03  & 90.44  &       & 12.77  & 83.98  &       & 16.69  & 64.26  &       & 44.62  & 24.88  \\
     \makebox[22.5mm][l]{Baseline + AugMix}~\cite{hendrycks_augmix_2020} & 8.86  & 92.31  &       & 12.46  & 85.12  &       & 16.07  & 66.56  &       & 38.76  & 28.25  \\
    \midrule
    Baseline + ScoreMix & 8.60  & 94.37  &       & 11.85  & 87.09  &       & 15.12  & 71.31  &       & 24.67  & 37.48  \\
    Baseline + $\text{ScoreMix}_{\text{ADA}}$ & 8.55  & \textbf{98.17}  &       & 11.78  & \textbf{89.69}  &       & 14.91  & 76.23  &       & 24.03  & 40.33  \\
    Baseline + $\text{ScoreMix}_{\text{LC}}$ & \textbf{8.42}  & 95.55  &       & \textbf{11.36}  & 87.14  &       & \textbf{14.43}  & \textbf{76.74}  &       & \textbf{22.51}  & \textbf{41.12}  \\
    \bottomrule
    \end{tabular}%
}
\end{table*}
}

\textcolor{black}{
We use FID~\cite{heusel2017gans} and Inception Score (IS) \cite{salimans_improved_2016} as quantitative metrics.
We sample generated samples with the truncation trick proposed in \cite{brock_large_2019}.
Similar to the unconditional image synthesis task, a lower FID indicates a better performance.
IS measures the diversity and quality of synthesized results using entropy. 
In this task, a higher IS score indicates a better result.
}

\vspace{2mm}
\textcolor{black}{
\noindent \textbf{Evaluation results.}
We evaluate our approaches against the mixing-based methods~\cite{zhang2017mixup,yun_cutmix_2019,hendrycks_augmix_2020}.
Table~\ref{tab:2.3.3.1} reports the FID and IS scores under 100\%, 50\%, 25\%, and 10\% data settings. 
The comparison results show that our approaches outperform the competing methods in terms of FID and IS.
The 10\% data setting only has fewer than 130 training samples for each class. 
Such scarce training data leads to degraded performance of the baseline method.
However, the proposed method significantly improves the model performance.
Similar to the results in unconditional image synthesis, combining the proposed method with the LC regularization~\cite{tseng_regularizing_2021} or adaptive discriminator augmentation method~\cite{karras_training_2020} brings further performance gain.
For instance, ScoreMix improves the FID score by 50.3\% (49.67 $\rightarrow$ 24.67), and $\text{ScoreMix}_{\text{LC}}$ further improves it by 8.8\% (24.67 $\rightarrow$ 22.51) under the 10\% data setting.
}

{
\captionsetup{labelfont={color=black},font={color=black}}
\begin{figure*}[t!]
\makebox[\textwidth][c]{\includegraphics[width=1\textwidth,height=0.55\textheight]{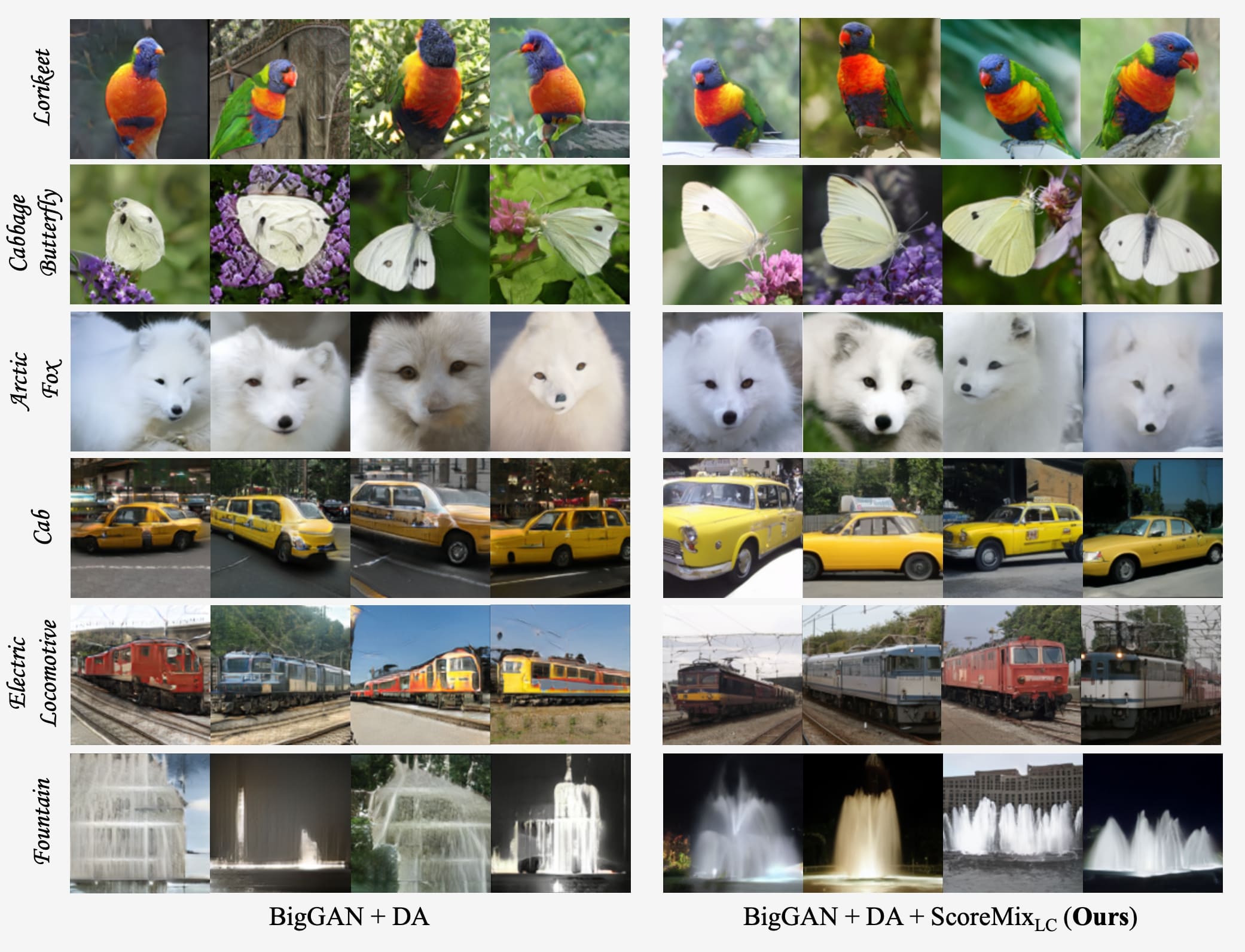}}
\caption{
Class-conditional ImageNet~\cite{russakovsky_imagenet_2015} samples generated under 10\% data setting. We use the BigGAN model~\cite{brock_large_2019} with the DiffAug scheme~\cite{zhao2020differentiable} as the baseline. Applying our approach to the baseline method synthesize more realistic results.
}
\label{fig:r1}
\end{figure*}
}

\textcolor{black}{
Figure~\ref{fig:r1} shows the synthesized samples from the baseline method and our approach.
These results show that our approach improves sample quality.
It can be observed that the samples from our approach have more plausible visual details than those from the baseline method (\eg, wings of the butterfly).
}

\vspace{3mm}

\subsection{Image-to-Image Translation}
Image-to-image translation (I2I) aims at learning the mapping from the source domain to the target domain. 
In the following, we perform two specific unsupervised I2I tasks in the data-limited regime.

\subsubsection{Animal Face Translation}

{
\renewcommand{\arraystretch}{1.3}
\begin{table*}[t!]
\begin{center}
\small
\caption{
Fr\'echet Inception Distance (FID, lower is better) and Learned Perceptual Image Patch Similarity (LPIPS, higher is better) of different methods on the AFHQ dataset~\cite{choi2020stargan}. ScoreMix is our proposed method. $\text{ScoreMix}_{\text{LC}}$ and $\text{ScoreMix}_{\text{ADA}}$ combine it with the regularization methods in \cite{tseng_regularizing_2021} and \cite{karras_training_2020}, respectively.
}
\label{tab:1}%
    \begin{tabular}{lccccccccccc}
    \toprule
    \multirow{3}[6]{*}{Method} & \multicolumn{5}{c}{Latent-guided translaion} &       & \multicolumn{5}{c}{Reference-guided translation} \\
\cmidrule{2-6}\cmidrule{8-12}          & \multicolumn{2}{c}{100\% data} &       & \multicolumn{2}{c}{10\% data} &       & \multicolumn{2}{c}{100\% data} &       & \multicolumn{2}{c}{10\% data} \\
\cmidrule{2-3}\cmidrule{5-6}\cmidrule{8-9}\cmidrule{11-12}          &~FID$\downarrow$~&~LPIPS$\uparrow$~& & FID\,$\downarrow$~&LPIPS\,$\uparrow$~&~& FID\,$\downarrow$~&~LPIPS\,$\uparrow$~& &~FID\,$\downarrow$~&~LPIPS\,$\uparrow$~\\
    \midrule
    Baseline: StarGAN v2~~\cite{choi2020stargan} & 16.18  & 0.450 & & 45.12 & 0.428 & & 19.78 & 0.432  &  & 39.23 & 0.407 \\
    \midrule
    \makebox[26mm][l]{Baseline \textbf{+} mixup}~\cite{zhang2017mixup} & 16.02 & 0.454 & & 28.15 & 0.450 &  & 18.51 & 0.453 & & 28.12 & 0.439  \\
    \makebox[26mm][l]{Baseline \textbf{+} CutMix}~\cite{yun_cutmix_2019} & 15.98 & 0.453 & & 41.36 & 0.444 & & 17.24 & 0.449 & & 26.33 & 0.442 \\
    \makebox[26mm][l]{Baseline \textbf{+} AugMix}~\cite{hendrycks_augmix_2020} & 15.72 & 0.443 & & 41.36 & 0.425 & & 16.98 & 0.434 & & 23.80 & 0.411 \\
    \makebox[27mm][l]{Baseline + \textcolor{black}{DiffAug}}~\cite{zhao2020differentiable} & 15.79 & 0.468 & & 25.11 & 0.460 &       & 16.89 & 0.477 &  & 23.02 & 0.459 \\
    \midrule
    \makebox[26mm][l]{Baseline + ReMix}~\cite{cao_remix_2021} & 15.26 & 0.482 & & 21.82 & 0.472 &       & 16.03 & 0.480 &  & 22.92 & 0.458 \\
    Baseline \textbf{+} ScoreMix & 14.03 & 0.470 & & 20.49 & 0.464 &       & 15.19 & 0.484 &  & 21.51 & 0.460 \\
    Baseline \textbf{+} \textcolor{black}{$\text{ScoreMix}_{\text{ADA}}$} & 14.18 & 0.479 & & 19.20 & 0.469 &       & 14.83 & 0.485 &  & 20.14 & 0.461 \\
    Baseline \textbf{+} \textcolor{black}{$\text{ScoreMix}_{\text{LC}}$} & \textbf{13.16} & \textbf{0.482} & & \textbf{18.83} & \textbf{0.474} &       & \textbf{14.11} & \textbf{0.485} &  & \textbf{17.99} & \textbf{0.466} \\
    \bottomrule
    \end{tabular}%
\end{center}
\end{table*}%
}

{
\renewcommand{\arraystretch}{1.4}
\captionsetup{labelfont={color=black},font={color=black}}
\setlength\tabcolsep{3pt}
\begin{table*}[t!]
\centering
\caption{
FID~\cite{heusel2017gans} (lower is better) and LPIPS~\cite{zhang2018unreasonable} (higher is better) as a function of training set size on the AFHQ~\cite{karras2020analyzing} dataset. ScoreMix is the proposed method. $\text{ScoreMix}_{\text{LC}}$ and $\text{ScoreMix}_{\text{ADA}}$ combine it with the regularization methods in \cite{tseng_regularizing_2021} and \cite{karras_training_2020}, respectively.
}
\label{tab:2.3.2.2}%
\makebox[\textwidth][c]{
    
    \begin{tabular}{lccccccccc}
    \toprule
    \multicolumn{1}{c}{\multirow{2}[4]{*}{Method}} & \multicolumn{4}{c}{Latent-guided translaion} &       & \multicolumn{4}{c}{Reference-guided translation} \\
\cmidrule{2-5}\cmidrule{7-10}          & 100\% data & 50\% data & 25\% data & 10\% data &       & 100\% data & 50\% data & 25\% data & 10\% data \\
    \midrule
    {\fontsize{9pt}{10pt}\selectfont Baseline}:~~~\cite{karras2020analyzing}\,\,+\,\cite{zhao2020differentiable} & 16.18 & 19.54 & 28.34 & 45.12 &       & 19.78 & 22.76 & 28.11 & 39.23 \\
    \midrule
    {\fontsize{9pt}{10pt}\selectfont Baseline + ScoreMix} & 14.03 & 15.42  & 18.76 & 20.49 &       & 15.19 & 17.09 & 18.30  & 21.51 \\
    {\fontsize{9pt}{10pt}\selectfont Baseline + $\text{ScoreMix}_{\text{ADA}}$} & 14.18 & 15.79 & 18.51 & 19.20  &       & 14.83 & 16.69 & 17.70  & 20.14 \\
    {\fontsize{9pt}{10pt}\selectfont Baseline + $\text{ScoreMix}_{\text{LC}}$} & 13.16 & 14.77 & 17.11 & 18.83 &       & 14.11 & 15.88 & 16.57 & 17.99 \\
    \bottomrule
    \end{tabular}%

}
\end{table*}%
}

\begin{figure*}[t!]
\begin{center}
\includegraphics[width=1.0\linewidth, height=275pt]{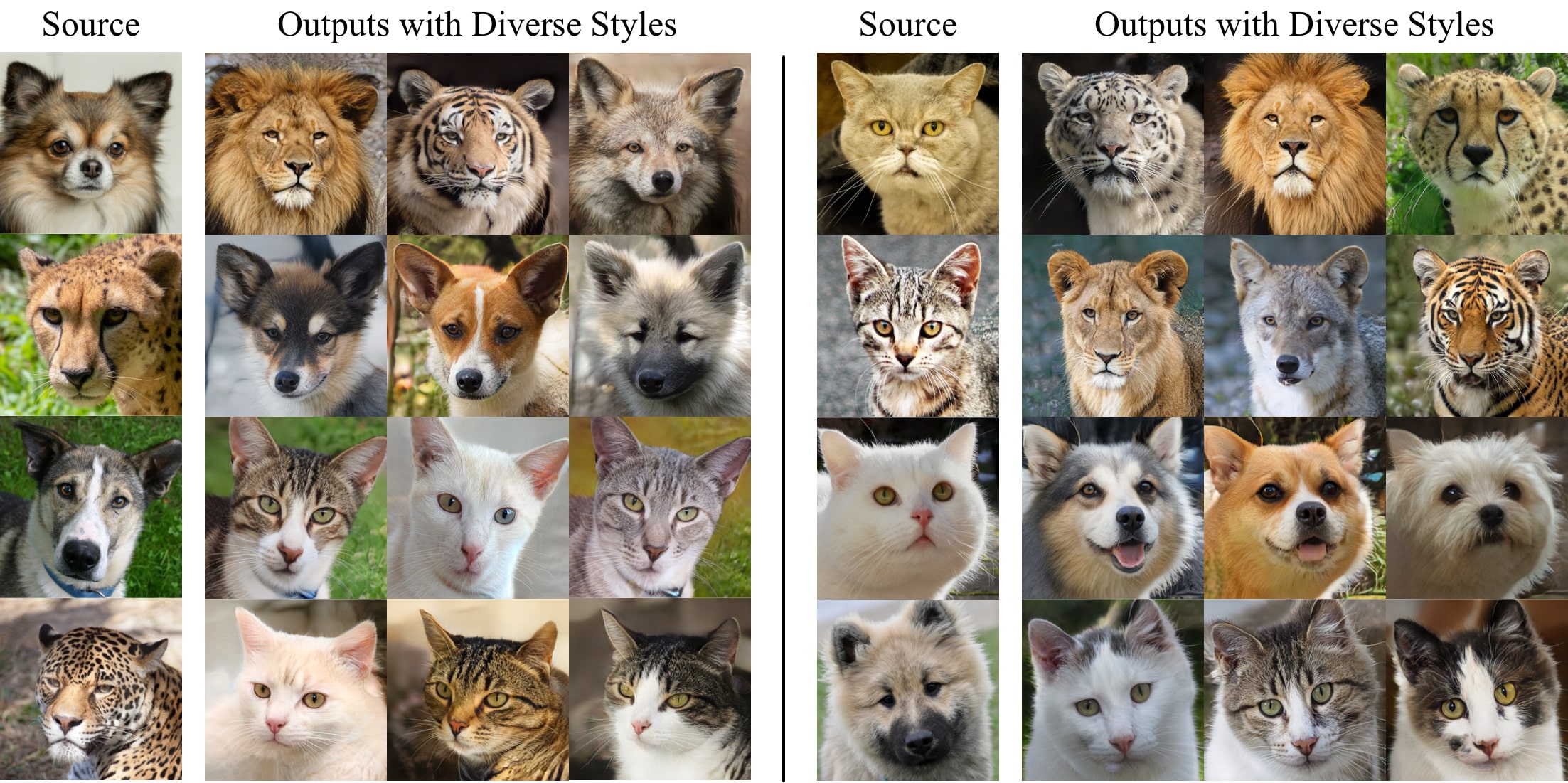}
\end{center}
\caption{
Diverse translation results on the AFHQ dataset~\cite{choi2020stargan}. 
Our model can learn to generate diverse high-quality results using only 10\% data in the training set.
}
\label{fig:afhq2}
\end{figure*}

\begin{figure*}[ht]
\begin{center}
\includegraphics[width=\linewidth]{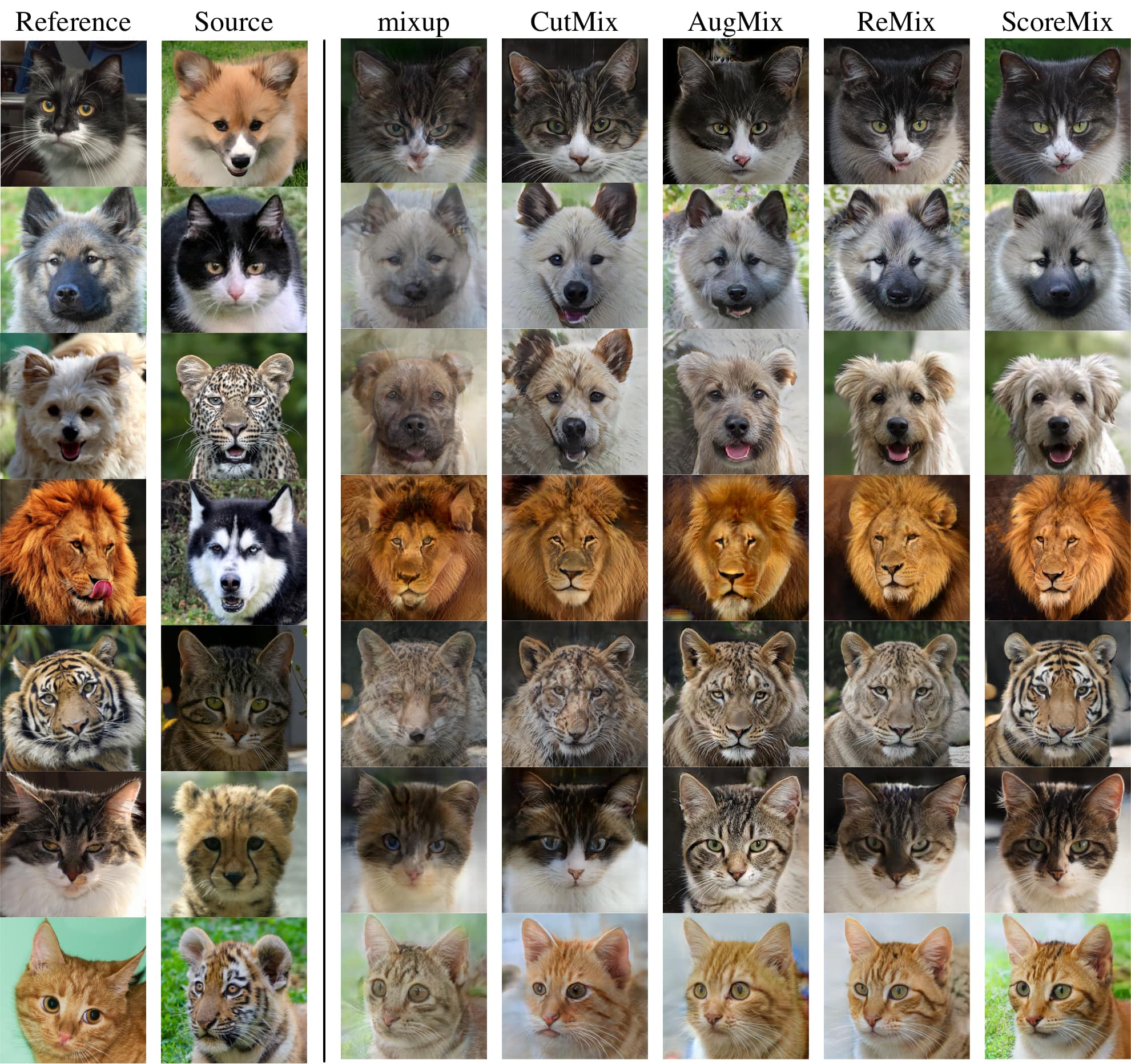}
\end{center}
\caption{
Visual samples generated by different methods under 10 \% data setting of the AFHQ dataset~\cite{choi2020stargan}. 
We use StarGAN v2~\cite{choi2020stargan} as the baseline. 
From left to right: the reference images, source images, baseline + mixup~\cite{zhang2017mixup}, baseline + CutMix~\cite{yun_cutmix_2019}, baseline + AugMix~\cite{hendrycks_augmix_2020}, baseline + ReMix~\cite{cao_remix_2021}, and baseline + ScoreMix (ours).
}
\label{fig:vis2}
\end{figure*}

In this subsection, we change the species of the input animal face (the source image). 
We consider the following two types of translations:
(1) reference-guided translation. Given a reference animal face (the reference image), we extract a style representation from it. 
Then, the generator mixes the style with the content of the source image, producing the translated result.
(2) latent-guided translation. Given a one-hot class label, we draw a latent code from a prior distribution conditioned on the label. 
The generator uses the latent code as the style representation. 
For this task, we use StarGAN v2~\cite{choi2020stargan} as the baseline method. 
It can perform both reference-guided translation and latent-guided translation. 
The training objective is minimizing the combination of the adversarial loss~\cite{goodfellow2014generative}, style reconstruction loss~\cite{zhu2017unpaired}, and style diversification loss~\cite{mao_mode_2019}. 
The resolutions of the input and output images are $256 \times 256$. 
To implement the baseline, we use the released official source codes\footnote{\url{https://github.com/clovaai/stargan-v2}}.

We evaluate the quality of synthesized images using FID~\cite{heusel2017gans} and Learned Perceptual Image Patch Similarity (LPIPS)~\cite{zhang2018unreasonable}.
The LPIPS score measures the diversity of images using the L1 distance in the feature space. 
In this task, the higher the LPIPS score, the better the synthesized results is.
We compute the FID and LPIPS scores for every pair of the image domains (\eg, dog$\rightarrow$cat, cat$\rightarrow$dog). 
Since the AFHQ dataset provides 3 different image domains, we compute these scores on 6 pairs of domains. 
Then, we report the average values of the scores as the evaluation metrics.

\vspace{2mm}
\noindent \textbf{Evaluation results.} 
Given a single source image, we generate diverse results by multiple random reference images.
Figure~\ref{fig:afhq2} shows diverse translation results by our method. 
We observe that the synthesized images are obviously different breeds of animals. That means our approach can generate distinctive styles.
In addition, the synthesized images preserve the content information such as pose and shape effectively.

We evaluate our method against existing mixing-based approaches, including the mixup~\cite{zhang2017mixup}, CutMix~\cite{yun_cutmix_2019}, AugMix~\cite{hendrycks_augmix_2020} and ReMix (our previous conference version~\cite{cao_remix_2021}) schemes. 
Figure~\ref{fig:vis2} shows some translation results. 
The synthesized images are supposed to preserve the low-level semantics (\eg, pose and shape) of the source image and the high-level semantics (\eg, color and fur) of the reference image.
The baseline model suffers from the overfitting problem and generates some unrealistic texture details.
Overall, the proposed method synthesizes images with higher visual quality than other schemes.

Table~\ref{tab:1} shows the FID and LPIPS scores of the evaluated methods. 
Our approach performs favorably against existing augmentation methods in terms of the quantitative metrics. 
The FID scores indicate that our results are more similar to the real data.
The LPIPS score of ScoreMix with 10\% data is even higher than that of the baseline with 100\% data.
These results demonstrate that the proposed method is effective for diverse and realistic image translation. 
\textcolor{black}{
Moreover, combining our approach with additional schemes~\cite{tseng_regularizing_2021,karras_training_2020} further improves model performance.
$\text{ScoreMix}_{\text{LC}}$ performs the best among all the competing methods.
}

\textcolor{black}{
Table~\ref{tab:2.3.2.2} presents the FID and LPIPS scores with different amounts of training data.
Under all the data settings, our approaches provide notable performance improvements.
Even for the 100\% data setting (about 14K samples), our approach is still particularly beneficial: ScoreMix improves the FID score by 13.3\% (16.18 $\rightarrow$ 14.03) on latent-guided translation and 23.2\% (19.78 $\rightarrow$ 15.19) on reference-guided translation.
}

\subsubsection{One-Shot Image-to-Image Translation}
In this subsection, we conduct I2I in the single image setting. 
Some recent methods~\cite{shaham_singan_2019,shocher_ingan_2019,park_contrastive_2020} have explored this challenging setting: only one source image and one target image are available for model training. 
Note that we do not use the pre-trained model on external data like the adaption-based I2I approaches~\cite{liu2019few,saito2020coco}. 

For this task, we use the official implementation\footnote{\url{https://github.com/taesungp/contrastive-unpaired-translation}} of the SinCUT method~\cite{park_contrastive_2020} as the baseline.
This method trains the I2I model in the patch-wise manner. 
Concretely, the input image is randomly cropped into 16 patches. 
Then, these patches are resized to $128 \times 128$ for model training. 
The translation model takes these patches as a batch of training data. 
The training is with the standard augmentation operations such as random scaling and horizontal flipping. 
In addition, we use the proposed ScoreMix method to increase the diversity of training image patches. 
During the test stage, the translation model takes the full resolution image as the input. 
This approach can translate a single image of arbitrary size. 

We use Single Image Fr\'echet Inception Distance (SIFID)~\cite{shaham_singan_2019} to evaluate the synthesized results. 
This metric calculates the FID score between the internal distributions of two images. 
We extract the features using the Inception-v3 network~\cite{szegedy2016rethinking}. 
Similar to the FID score, a lower SIFID score between the synthesized image and target image indicates a better result. 

\begin{figure*}[!t]
\begin{center}
\includegraphics[width=1.0\linewidth, height=310pt]{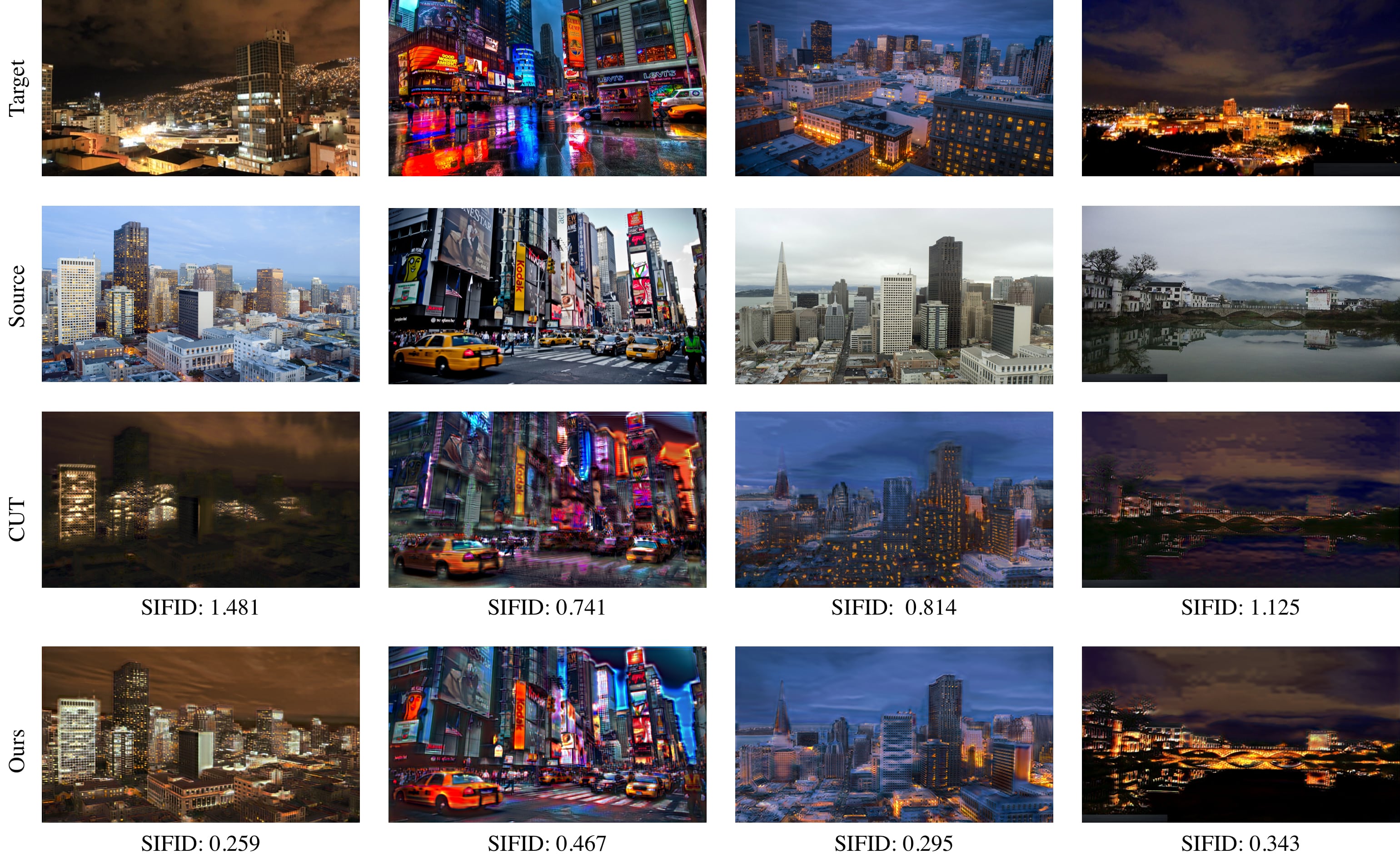}
\end{center}
\caption{
Image-to-image translation results from two unpaired training images. 
We train the translation model with \textit{one source image} and \textit{one target image}. 
The results are synthesized by the SinCUT method~\cite{park_contrastive_2020} and our approach. We report Single Image Fr\'echet Inception Distance~\cite{shaham_singan_2019} (SIFID, lower is better).}
\label{fig:si2i}
\end{figure*}

\vspace{1mm}
\noindent \textbf{Evaluation results.} Figure~\ref{fig:si2i} shows some translation results and the corresponding SIFID scores. We use the collected source and target images from Luan~\etal~\cite{luan_deep_2017}. The translated images are supposed to preserve the general structure of the source image and match the textures of the target image. Although this task is challenging due to the limited data, the proposed method achieves reasonable results. The SinCUT method~\cite{park_contrastive_2020} suffers from the overfitting problem and synthesizes distorted images. 
In contrast, the proposed method transfers the style of the target image more faithfully and shows fewer artifacts.
Moreover, the proposed method obtains significantly lower SIFID scores.
These results show that our synthesized images are more similar to the real data.
 
We conduct a user study to validate the proposed method. 
First, we use the SinCUT method~\cite{park_contrastive_2020} and our approach to perform a total of 60 one-shot image-to-image translation tasks. 
The source-target image pairs for training are publicly available\footnote{\url{https://github.com/luanfujun/deep-photo-styletransfer}}.
Each method synthesizes 60 translated results.
Then, we ask evaluators to vote for the synthesized images from the SinCUT method and our approach based on image quality.
We keep the translation methods anonymous in the evaluation.
The source and target images are provided for the evaluators as references.
We distribute questionnaires to online users and finally get valid feedbacks from 71 human evaluators. 
We did not collect any private information.

Table~\ref{tab:si2i} shows the competition results.
We compute the SIFID scores and record user preferences.
Then, we count the number of times each method performs better than its competitor and report the percentage in the table.
For each task, any method with the majority of the user votes is better in the human evaluations.
As for the SIFID score, lower is better.
The competition results show that the proposed method generally performs better regarding both quantitative evaluations and human evaluations.

{
 \aboverulesep=0ex
 \belowrulesep=0ex
 \renewcommand{\arraystretch}{2}
\begin{table}[t!]
\begin{center}
\small
\caption{
Our approach vs the SinCut method~\cite{park_contrastive_2020}. 
We evaluate the two methods on 60 different one-shot image-to-image translation tasks. 
The metrics include Single Image Fr\'echet Inception Distance~\cite{shaham_singan_2019} (SIFID) and human evaluation.
}
\label{tab:si2i}%
    \begin{tabular}{cccc}
    \toprule
    \multicolumn{2}{c}{\multirow{2}[4]{*}{Metrics}} & \multicolumn{2}{c}{Human Evaluation} \\
\cmidrule{3-4}    \multicolumn{2}{c}{} &  Ours $>$ SinCut &  Ours $<$ SinCut \\
    \midrule
    \multirow{2}[4]{*}{\rotatebox{90}{~~SIFID}} & Ours $>$ SinCut & 83.3\% & 6.6\% \\
\cmidrule{2-4}          & Ours $<$ SinCut & 1.7\%    & 8.4\% \\
    \bottomrule
    \end{tabular}%
\end{center}
\end{table}%
}

\section{Discussion}
In this work, we have verified that combining the proposed method with additional augmentation operations, including \cite{zhao2020differentiable,lecun_gradient-based_1998,krizhevsky_imagenet_2012,devries2017improved,karras_training_2020,tseng_regularizing_2021}, further improves model performance. 
Incorporating recent methods that are orthogonal to this work, \eg, the RL-based augmentations~\cite{cubuk_autoaugment_2019,lim_fast_2019}, may perform well. 
In addition, the proposed method could be easily used in the semi-supervised learning framework~\cite{li_dividemix_2020,sohn_fixmatch_2020,berthelot2019mixmatch}. We leave the study to future work.

The focus of this paper is how to generate images plausible to human visual perception.
We have shown that the proposed data augmentation method improves the sample quality using automatic metrics~\cite{heusel2017gans,zhang2018unreasonable,shaham_singan_2019,binkowski_demystifying_2018} and subjective evaluation.
However, whether or not the proposed method is beneficial still depends on the applications. 
It might be less suited for some particular applications, such as medical imaging~\cite{guibas2017synthetic} and biology~\cite{costa2017end}.

\section{Conclusion}
This paper introduces a mixing-based data augmentation method to tackle the overfitting problem of GANs.
We propose to optimize the augmented samples by minimizing the norms of the data scores.
This approach prevents manifold intrusion and produces high-quality augmented samples.
Hence, it increases the diversity of image samples, facilitating GAN training in image synthesis.
In addition, the proposed approach is scalable to various image synthesis tasks and requires no data-specific fine-tuning.
We demonstrate that our method vastly improves image quality and quantitative metrics in numerous tasks, especially when the training data is limited.

\ifCLASSOPTIONcaptionsoff
  \newpage
\fi



\end{document}